\documentclass{article}

\usepackage[preprint]{neurips_2024}

\usepackage[T1]{fontenc}    
\usepackage{pst-poker}
\usepackage{hyperref}       
\usepackage{url}            
\usepackage{booktabs}       
\usepackage{amsfonts}       
\usepackage{nicefrac}       
\usepackage{microtype}      
\usepackage{xcolor} 
\usepackage{amsmath}
\usepackage{tikz}
\usepackage{multirow}
\usepackage{graphicx} 
 \usepackage{grffile}
\usepackage[capitalise]{cleveref}
\usepackage{amssymb}
\usepackage{subfigure}
\usepackage[numbers]{natbib} 
\title{Spectral Informed Neural Network: An Efficient and Low-Memory PINN}

\author{%
  Tianchi Yu\thanks{Equal contribution} \\
  Skolkovo Institute \\ of Science and Technology\\
  \texttt{Tianchi.Yu@skoltech.ru} \\
  \And
  Yiming Qi\footnotemark[1] \\
  Peking University \\
  \texttt{2001111682@stu.pku.edu.cn} \\
  \AND
  Ivan Oseledets\\
  AIRI;\\
  Skolkovo Institute \\ of Science and Technology\\
  \texttt{I.Oseledets@skoltech.ru} \\
  \And
  Shiyi Chen \\
  Peking University;\\ 
  Southern University \\ of Science and Technology\\
  \texttt{chensy@sustech.edu.cn} \\
}

\begin{document}
\maketitle
\begin{abstract}
With growing investigations into solving partial differential equations by physics-informed neural networks (PINNs), more accurate and efficient PINNs are required to meet the practical demands of scientific computing. One bottleneck of current PINNs is computing the high-order derivatives via automatic differentiation which often necessitates substantial computing resources. In this paper, we focus on removing the automatic differentiation of the spatial derivatives and propose a spectral-based neural network that substitutes the differential operator with a multiplication. Compared to the PINNs, our approach requires lower memory and shorter training time. Thanks to the exponential convergence of the spectral basis, our approach is more accurate. Moreover, to handle the different situations between physics domain and spectral domain, we provide two strategies to train networks by their spectral information. Through a series of comprehensive experiments, We validate the aforementioned merits of our proposed network. 
\end{abstract}
\section{Introduction}
With the rapid advancements in machine learning and its related theories,
integrating mathematical models with neural networks provides a novel framework for embedding physical laws in scientific research. The representative methods are the Physics-Informed Neural Networks (PINNs)~\cite{raissi2019physics} and the Deep Ritz method~\cite{yu2018deep}. PINNs have garnered significant attention because of their ability of solving partial differential equations (PDEs) without suffering from the curse of dimensionality (CoD)~\cite{wojtowytsch2020can,han2018solving}, thanks to the Monte Carlo method~\cite{rubinstein2016simulation}, automatic differentiation (AD)~\cite{baydin2018automatic}, and universal approximation theorem~\cite{hornik1991approximation}. Additionally, PINNs demonstrate the merits in handling imperfect data~\cite{karniadakis2021physics}, extrapolation~\cite{yang2021b,ren2022phycrnet}, and interpolation~\cite{sliwinski2023mean}. These capabilities have propelled PINNs into a wide applications, including but not limited to fluid dynamics~\cite{jin2021nsfnets}, aerodynamics~\cite{mao2020physics}, surface physics~\cite{fang2019physics}, power systems~\cite{misyris2020physics}, and heat transfer~\cite{gao2021phygeonet}.

However, using AD to compute the loss function (mainly the residual term) during training, is inefficient and computational expensive. To address this problem, various approaches have been proposed that leverage alternative numerical methods to replace AD. For instance, PhyCRNet~\cite{ren2022phycrnet,rao2023encoding} utilizes the finite difference method (FDM) to replace AD; DTPINN~\cite{sharma2022accelerated} applies the finite difference for radial basis functions instead of high-order derivatives; sPINN~\cite{xia2023spectrally} employs the spectral method of orthogonal polynomials to avoid computing the derivatives and Spectral PINNs~\cite{lutjens2021spectral} utilize Hermite polynomials for uncertainty quantification. 
On the other hand, some researchers focus on improving the efficiency of AD, the representative method is the Taylor-mode~\cite{griewank2008evaluating}.

Unfortunately, directly embedding those conventional numerical methods into PINNs will resuffer the CoD. However, the spectral method offers a promising alternative by approximating solutions within the spectral domain. By mapping sampled frequencies to their corresponding coefficients, the spectral method can be embedded into PINNs in a manner that circumvents the CoD. sPINNs have utilized a similar idea in their implementation of the spectral method; nevertheless, they only illustrate a few experiments of linear PDEs on finite, semi-infinite, and infinite intervals with truncated Chebyshev polynomials, Laguerre polynomials, and Hermite polynomials, respectively. Technically speaking, when considering non-linear PDEs, aliasing errors caused by nonlinear terms should be suppressed in the spectral domain via dealiasing operations. While in sPINNs, they didn't discuss it. Furthermore, their input is the spatial and temporal grid points,  which makes sPINNs unable to fully exploit the mesh-free features of spectral methods and adaptively sample from the frequencies.

In this paper, we propose the Spectral-Informed Neural Networks (SINNs) as an efficient and low-memory approach to train networks for the PDEs with periodic boundary conditions by the spectral information based on Fourier basis. Compared to PINNs, our SINNs utilize a precise and efficient alternative to AD to acquire spatial derivatives; the input is the frequencies of the Fourier basis rather than the grid points from the physics domain, and the output is the coefficients in the spectral domain rather than the physical solution. Furthermore, the merit of the exponential convergence for approaching any smooth function~\cite{Canuto1988Spectral} not only makes spectral methods achieved superior accuracy in traditional computational science~\cite{Canuto1988Spectral, Orszag1971}, but it also enables SINNs to achieve smaller errors.

Our specific contributions can be summarized as follows:
\begin{enumerate}
    \item We propose a method that tackles the spatial derivatives of the network to deal with the high GPU memory consumption of PINNs.
    \item We propose two strategies to approximate more of the primary features in the spectral domain by learning the low-frequency preferentially to handle the difference between SINNs and PINNs.
    \item Our experiments show that the method can reduce the training time and improve the accuracy simultaneously.
\end{enumerate}

The paper is structured as follows: In \cref{sec:PINNs}, we provide a concise overview of PINNs and discuss AD and its developments. Using a simple experiment, we highlight the challenge encountered in computing high-order derivatives within PINNs. To address this drawback, in \cref{sec:SINNs}, we propose our SINNs and give an intuitive understanding with two examples. In \cref{sec:Experiment}, we demonstrate state-of-the-art results across a comprehensive experiment. Finally, \cref{section: conlclusion} provides a summary of our main research and touches upon remaining limitations and directions for future research.

\section{Physics-informed neural networks (PINNs)}\label{sec:PINNs}
We briefly review the physics-informed neural networks (PINNs)~\cite{raissi2019physics} in the context of
inferring the solutions of PDEs. Generally, we consider PDEs for $\boldsymbol{u}$ taking the form
\begin{equation}
\begin{aligned}
    & \partial_{t}\boldsymbol{u}+\mathcal{N}[\boldsymbol{u}]=0, \quad t \in[0, T],\ \boldsymbol{x} \in \Omega, \\
    & \boldsymbol{u}(0, \boldsymbol{x})=\boldsymbol{g}(\boldsymbol{x}), \quad \boldsymbol{x} \in \Omega, \\
    & \mathcal{B}[\boldsymbol{u}]=0, \quad t \in[0, T],\ \boldsymbol{x} \in \partial \Omega,
\end{aligned}\label{PDE}
\end{equation}
where $\mathcal{N}$ is the differential operator, $\Omega$ is the domain of grid points, and $\mathcal{B}$ is the boundary operator. 

The ambition of PINNs is to obtain the unknown solution $\boldsymbol{u}$ to the PDE system \cref{PDE}, by a neural network $\boldsymbol{u}^{\theta}$, where $\theta$ denotes the parameters of the neural network. The constructed loss function is:
\begin{equation}\label{PINN}
\mathcal{L}(\theta)=\mathcal{L}_{i c}(\theta)+\mathcal{L}_{b c}(\theta)+\mathcal{L}_r(\theta) ,
\end{equation}
where
\begin{equation}
\begin{aligned}
& \mathcal{L}_r(\theta)=\frac{1}{N_r} \sum_{i=1}^{N_r}\left|\partial_{t}\boldsymbol{u}^\theta\left(t_r^i, \boldsymbol{x}_r^i\right)+\mathcal{N}\left[\boldsymbol{u}^\theta\right]\left(t_r^i, \boldsymbol{x}_r^i\right)\right|^2,\\
& \mathcal{L}_{i c}(\theta)=\frac{1}{N_{i c}} \sum_{i=1}^{N_{i c}}\left|\boldsymbol{u}^\theta\left(0, \boldsymbol{x}_{i c}^i\right)-\boldsymbol{g}\left(\boldsymbol{x}_{i c}^i\right)\right|^2, \\
& \mathcal{L}_{b c}(\theta)=\frac{1}{N_{b c}} \sum_{i=1}^{N_{b c}}\left|\mathcal{B}\left[\boldsymbol{u}^\theta\right]\left(t_{b c}^i, \boldsymbol{x}_{b c}^i\right)\right|^2, \\
\end{aligned}
\end{equation}
corresponds to the three equations in \cref{PDE} individually; $\boldsymbol{x}_{i c}^i,\boldsymbol{x}_{b c}^i,\boldsymbol{x}_{r}^i$ are the sampled points from initial constraint, the boundary constraint, and the residual constraint; $N_{i c},N_{b c},N_{r}$ are the total number of sampled points correspondingly.
\subsection{Automatic differentiation (AD)}\label{section: AD}
\begin{figure}[!htbp]
  \centering 
  \subfigure[\label{memory_test_1} Allocated memory of the GPU]{\includegraphics[width=0.329\linewidth]{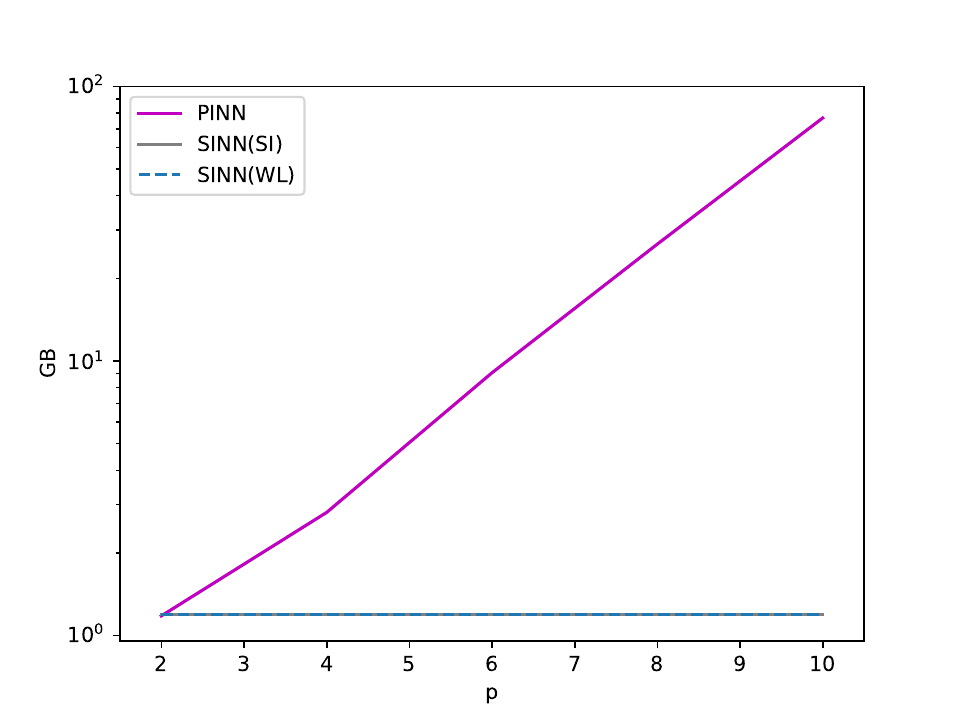}} 
  \subfigure[\label{memory_test_2} Relative error for $p=2$]{\includegraphics[width=0.329\linewidth]{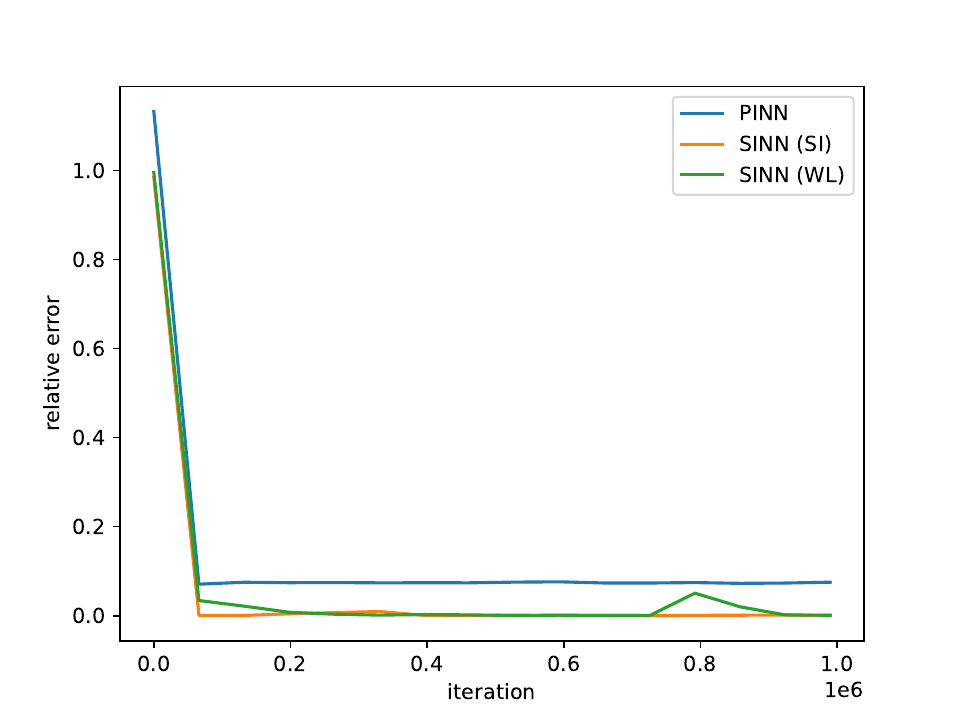}}
  \subfigure[\label{memory_test_3} Relative error for $p=6$]{\includegraphics[width=0.329\linewidth]{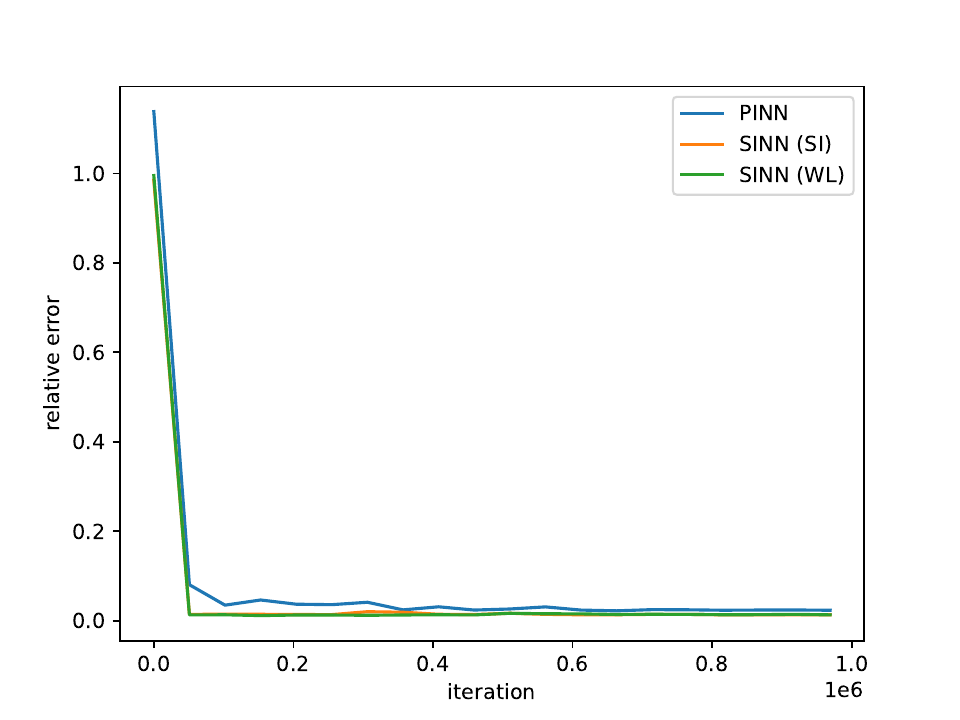}}
  \caption{In the experiments for \cref{eq: highorder} with different $p$, \subref{memory_test_1} depicts the measured maximum memory of the GPU allocated during training, implying that the memory increases exponentially with $p$ for the spatial derivative $\partial^p u/\partial x^p$ in PINNs and is constant for the multiplication $k^p \hat{u}$ in SINNs. And the corresponding relative L2 errors with iterations for $u(t=0.1,x)$ are shown in \subref{memory_test_2} and \subref{memory_test_3} for $p=2$, $p=6$ respectively, indicating that SINNs are more accurate than PINNs whatever dealing with low-order derivatives or high-order derivatives. The further discussion of the experiments is presented in \cref{sec:highorder}.}
  \label{fig:memory}
\end{figure}

AD gives the required derivative of an overall function by combining the derivatives of the constituent operations through the chain rule based on evaluation traces. Herein, AD is used to calculate the derivatives respect to the input points in PINNs. However, AD demands both memory and computation that scale exponentially with the order of derivatives (\cref{memory_test_1}) although there are investigations~ \cite{griewank2008evaluating,bettencourt2019taylor,tan2023higher} on computing high-order derivatives efficiently by Faà di Bruno's formula\footnote{the further discussion is in \cref{Appendix: FB_formula}}. Alternatively, replacing the high-order derivatives by simple multiplication, SINNs can reduce both memory and training time (\cref{memory_test_1} and \cref{table: differenorder}).

\begin{figure}[!htbp]
  \centering   
  \subfigure[PINNs] {\centering     
\includegraphics[width=0.8\columnwidth]{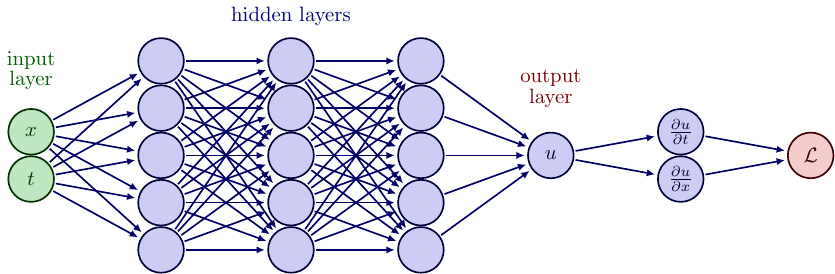}  
}\\     
  \subfigure[SINNs] { \centering     
\includegraphics[width=0.8\columnwidth]{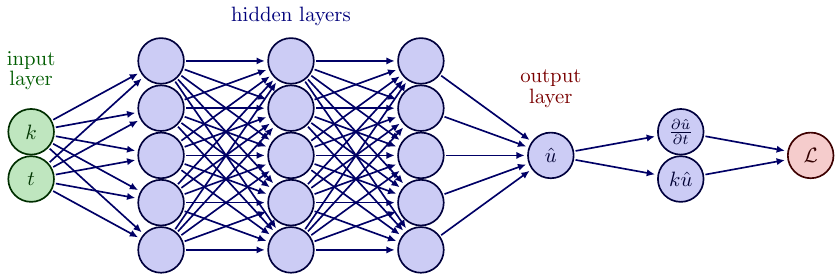}     
}        
\caption{Comparison of PINNs and SINNs. In PINNs, the input is the spatial and temporal grid points from the domain, the output is the physical solution $u$. In SINNs, the input is the frequencies of the Fourier basis and the temporal grid points from the domain, the output is the coefficients $\hat{u}$ in the spectral domain.}\label{fig: Architecture}
\end{figure}
\section{Spectral Information Neural Networks (SINNs)}\label{sec:SINNs}
To implement spectral method into PINNs, the Fourier operator $\mathcal{F}$ is applied to \cref{PDE}, converting the solution from physics domain to the frequency domain. 
Practically, we use $\mathcal{F}_N$ to represent the $N$-th truncated Fourier operator:
\begin{equation}
    \mathcal{F}_N[u]\left(t,x\right)=\sum_{k=-N/2}^{N/2-1} \hat{u}\left(t,k\right)e^{ikx} ,
\end{equation}
  where $\hat{u}$ corresponds to the Fourier coefficients of $u$, and $i\equiv \sqrt{-1}$ is the unit imaginary number.
One can easily obtain the following derivatives:
\begin{equation}
    \partial_t \mathcal{F}_N[u]\left(t,x\right)=\sum_{k=-N/2}^{N/2-1} \partial_t\hat{u}\left(t,k\right)e^{ikx},
\end{equation}
\begin{equation}
    \partial_{x} \mathcal{F}_N[u]\left(t,x\right)=\sum_{k=-N/2}^{N/2-1} ik\hat{u}\left(t,k\right)e^{ikx}. \label{eq: vanish}
\end{equation}
\Cref{eq: vanish} straightforwardly shows the vanishment of the AD for spacial derivatives in spectral domain.
To demonstrate how SINNs work intuitively, we study two representative examples, including a two-dimensional (2-D) heat equation (linear) and a 2-D Navier-Stokes (NS) equation (non-linear).

\subsection{Physical model for heat transfer}\label{section: linear}
Ideally, the heat transfer can be described by the heat equation which is investigated widely in mathematics as one of the prototypical PDEs. Given $\Omega \subset \mathbb{R}^2 $, consider the 2-D heat equation with periodic boundary condition: 
\begin{equation}
\begin{gathered}
\partial_{t}u(t,\boldsymbol{x})=\partial_{xx}u(t,\boldsymbol{x})+\partial_{yy}u(t,\boldsymbol{x}), \quad t \in[0, T],\ \boldsymbol{x} \in \Omega,\\
    u(0,\boldsymbol{x})=\boldsymbol{g}(\boldsymbol{x}), \quad \boldsymbol{x} \in \Omega .
\end{gathered}
\end{equation}
When applying on the loss function\cref{PINN}, the residual loss $\mathcal{L}_r(\theta)$ is explicitly expressed by:
\begin{equation}
\mathcal{L}_r(\theta)=\frac{1}{N_r} \sum_{i=1}^{N_r}\left|\partial_{t}u^\theta\left(t_r^i, \boldsymbol{x}_r^i\right)-\partial_{xx}u^\theta\left(t_r^i, \boldsymbol{x}_r^i\right)-\partial_{yy}u^\theta\left(t_r^i, \boldsymbol{x}_r^i\right)\right|^2,
\end{equation}

In our SINNs, the loss function is in spectral domain without the boundary constraint due to the periodic nature of the Fourier basis and the loss function $\mathcal{L}\left(\theta\right)$ is modified to:
\begin{equation}\label{SINN-HeatEq}
\Tilde{\mathcal{L}}(\theta)=\Tilde{\mathcal{L}}_{i c}(\theta)+\Tilde{\mathcal{L}}_r(\theta) ,
\end{equation}
where
\begin{equation}
\begin{aligned}
\Tilde{\mathcal{L}}_r(\theta)=&\frac{1}{N_r} \sum_{i=1}^{N_r}\left|\partial_{t}\hat{u}^\theta\left(t_r^i, \boldsymbol{k}^i\right)+\left(k_x^i\right)^2\hat{u}^\theta\left(t_r^i, \boldsymbol{k}^i\right)+\left(k_y^i\right)^2\hat{u}^\theta\left(t_r^i, \boldsymbol{k}^i\right)\right|^2,\\
\Tilde{\mathcal{L}}_{i c}(\theta)=&\frac{1}{N_{i c}} \sum_{i=1}^{N_{i c}}\left|\hat{u}^\theta\left(0, \boldsymbol{k}^i\right)-\boldsymbol{g}\left(\boldsymbol{k}^i\right)\right|^2, \\
\end{aligned}
\end{equation}
and $\boldsymbol{k}^i=(k_x^i,k_y^i) \in \left[-{N/ 2},{N/ 2}-1 \right]^2$ is the sampled frequency from spectral domain.

\subsection{Physical model for incompressible flows}\label{section: IncomFlows}
The mathematical model describing incompressible flows is the incompressible NS equation, namely,
\begin{subequations}\label{eq:ns}\begin{equation}\label{eq:ns_continuity}
\nabla \cdot \boldsymbol{u}=0, \quad t\in [0,T], \ \boldsymbol{x} \in \Omega,
\end{equation}\begin{equation}\label{eq:ns_momentum}
\partial_t \boldsymbol{u}+\boldsymbol{u} \cdot \nabla \boldsymbol{u}=-\nabla p+\nu\triangle \boldsymbol{u},\quad t\in [0,T], \ \boldsymbol{x} \in \Omega,
\end{equation}
\begin{equation}
    \boldsymbol{u}(0,\boldsymbol{x})=\boldsymbol{\boldsymbol{g}}(\boldsymbol{x}), \quad \boldsymbol{x} \in \Omega,
\end{equation}
\end{subequations}
where $\nabla = (\partial_x, \partial_y)$ is the gradient operator, $\boldsymbol{u}(t,\boldsymbol{x}) =(u,v)$ is the hydrodynamic velocity, $p(t,\boldsymbol{x})$ is the mechanical pressure, $\triangle = \nabla\cdot\nabla$ is the Laplace operator, and $\nu$ is kinematic viscosity.
By Fourier transform and appropriate derivation tricks (see \cref{Appendix: detailed NS}), the continuity equation~\cref{eq:ns_continuity} and momentum equation~\cref{eq:ns_momentum} can be expressed in the spectral domain:
\begin{subequations}\label{PS_NS}\begin{equation} 
\boldsymbol{k} \cdot \hat{ \boldsymbol{u} } = 0 , 
  \label{PS_mass}
\end{equation}
\begin{equation} 
\partial_t \hat{\boldsymbol{u}} = -\left(1- \frac{\boldsymbol{k}\boldsymbol{k}\cdot}{ |\boldsymbol{k}|^2} \right) \hat{\boldsymbol{N}} - \nu |\boldsymbol{k}|^2 \hat{\boldsymbol{u}},
  \label{PS_momentum}
\end{equation}\end{subequations}
  where $|\boldsymbol{k}|^2=\boldsymbol{k}\cdot\boldsymbol{k}$ is the inner product for $\boldsymbol{k}=(k_x,k_y)$, and $\hat{\boldsymbol{N}}$ is the non-linear term in the spectral domain, which has the rotational form $\boldsymbol{N} = (\boldsymbol{\nabla}\times\boldsymbol{u})\times \boldsymbol{u}$ in the physical space~\cite{Canuto1988Spectral}.
The continuity equation \cref{PS_mass} can be preserved strictly by the projection in our SINNs\footnote{There are many approaches to do such kind of projection, in this paper, we choose $\hat{\boldsymbol{u}}=\left(1-\frac{\boldsymbol{k} \boldsymbol{k}}{\boldsymbol{k}^2} \cdot\right) \hat{\boldsymbol{u}}$.}; however, the non-linear term has the challenge of dealing with the aliasing error, which will be further discussed in \cref{Appendix: aliasing}. 
After solving the aliasing error, the loss function is
\begin{small}
\begin{equation}\label{SINN-NSEq} 
    \Tilde{\mathcal{L}}_r(\theta)=\frac{1}{N_r} \sum_{i=1}^{N_r}\left|\mathcal{F}^{-1}[\partial_t \hat{\boldsymbol{u}}\left(t_r^i, \boldsymbol{k}\right)]+\mathcal{F}^{-1}\left[\left(1- \frac{\boldsymbol{k}\boldsymbol{k}\cdot }{ |\boldsymbol{k}|^2}\right) \hat{\boldsymbol{N}}\left(t_r^i, \boldsymbol{k}\right)\right]+\nu \mathcal{F}^{-1}\left[ |\boldsymbol{k}|^2\hat{\boldsymbol{u}}\left(t_r^i, \boldsymbol{k}\right) \right] \right|^2.
\end{equation}
\end{small}

\subsection{Importance optimisation}\label{sec: Importrance}
Compared to PINNs, the main divergence is the importance of different input points. 
Although the literature on sampling method ~\cite{tang2023pinns,wu2023comprehensive,lu2021deepxde} shows that the importance of the input points in physics domain can be dependent on the corresponding residual, generally speaking, every point is equally important without any prior knowledge. But for SINNs, normally the importance decreases as the corresponding frequency increases.
For instance, the energy spectrum in the inertial ranges of 2-D turbulence (described by the 2-D NS equation) satisfies the scaling relation~\citep{Kraichnan1967}:
\begin{equation}
\sum\limits_{n- {\frac{1}{2}}\leq |\boldsymbol{k}|< n+{\frac{1}{2}}} |\hat{\boldsymbol{u}}\left(t,\boldsymbol{k}\right)|^2
  \sim n^{-3}.
\end{equation}
Similar physical analysis can be performed for 1-D problems, and the physical background demonstrates that the Fourier coefficient $\hat{\boldsymbol{u}}$ decreases rapidly as an increase in frequency due to the effect of viscosity.
In practical, comprehensive experiments in Fourier Neural Operator~\cite{li2020fourier} show that the truncated Fourier modes can contain most features of the PDEs. Herein, PINNs learn every input point equally, while SINNs are supposed to learn the low-frequency points preferentially. To train the network based on the aforementioned divergence, we propose two strategies:

\textbf{Sampling by importance (SI)}:  
Suppose $p(\boldsymbol{k})$ is the probability
density function (PDF) used to sample the residual points, then the $p(\boldsymbol{k})$ in sampling by importance is defined as:
\begin{equation}
    p(\boldsymbol{k}) \propto \frac{1}{\|\boldsymbol{k}\|_1+\beta}+\alpha,
\end{equation}
where $\alpha,\beta$ are two hyperparameters. This PDF makes us sample more points on the low frequencies and less points on the high frequencies.

\textbf{Weighted residual loss (WL)}: However, although the capability of the neural network is hard to estimate, we should exploit it as much as possible. In SI, training SINNs with high-frequencies is in pretty low possibilities, such kind of strategy makes network can't learn high-frequencies even if the constructed network can learn both low-frequencies and high-frequencies. To address this dilemma, inspired by the CausalPINN\cite{wang2022respecting} and taking the 2-D heat equation as an example, we propose the weighted loss function\footnote{Although the weighted residual loss looks complicated, it can be simplified (See  \cref{Appendix: Weighted loss})}:
\begin{equation}
    \Tilde{\mathcal{L}}_r(\theta)=\frac{1}{M}\sum_{i=0}^{M}\exp\left(-\epsilon\sum_{k=0}^{i-1}\mathcal{L}_r^k\right)\mathcal{L}_r^i\label{eq:cauloss_c} ,
\end{equation}
where 
\begin{equation}
    \mathcal{L}_r^i=\sum_{\|\boldsymbol{k}\|_1 = i} \left|\partial_{t}\hat{u}^\theta\left(t_r, \boldsymbol{k}\right)
      +\left(k_x\right)^2\hat{u}^\theta\left(t_r, \boldsymbol{k}\right)
      +\left(k_y\right)^2\hat{u}^\theta\left(t_r, \boldsymbol{k}\right)\right|^2,
    \label{fft_loss}
\end{equation}
$\epsilon$ is the hyperparameter, and $M=\max\left(\|\boldsymbol{k}\|_1\right)$. Note that when solving on non-linear PDEs, the inverse operator requires the accurate coefficients of frequencies but the WL gives a different weight on different frequency. Herein WL can't be applied on non-linear PDEs.

\subsection{Spectral convergence}\label{section: spectral convergence}
Regardless of the convergence analysis in temporal domain\footnote{Generally, the temporal error is always much smaller than spatial error so we can ignore the temporal error.},assume that the capability of MLP is powerful enough, $u \in C^{\infty}(\Omega,\mathbb{R})$ is a smooth function from a subset $\Omega$ of a Euclidean $\mathbb{R}$ space to a Euclidean space $\mathbb{R}$, and $N$ is the number of discretized points.

Firstly, let's review the convergence rate of PINNs. Suppose $u^*$ is the exact solution in the domain $\Omega$ and
\begin{equation}
\begin{gathered}
    \theta^*\triangleq \arg \min_\theta \int_\Omega \mathcal{L}_r\left[u^\theta(x)\right] \mathrm{d}x, \\
    \theta^*_N\triangleq \arg\min_\theta \sum_{i=1}^{N} \mathcal{L}_r\left[u^\theta(x_i)\right] .
\end{gathered}
\end{equation}
Then
\begin{equation}\label{eq: mc}
    \|u^{\theta^*_N}-u^*\|_\Omega \leq  \underset{\text{statistical error}}{\|u^{\theta^*_N}- u^{\theta^*}\|_\Omega}+\underset{\text{approximation error}}{\| u^{\theta^*}-u^*\|_\Omega},
\end{equation}
where approximation error depends on the capability of PINNs. As the capability of MLP is powerful enough, $\| u^{\theta^*}-u^*\|_\Omega \ll \|u^{\theta^*_N}- u^{\theta^*}\|_\Omega$. Additionally, based on the Monte Carlo method, the statistical error is $\mathcal{O}\left(N^{-1/2}\right)$ \cite{quarteroni2006numerical}, then:
\begin{equation}
    \|u^{\theta^*_N}-u^*\|_\Omega  = \mathcal{O}\left(N^{-1/2}\right).
\end{equation}

As for SINNs, with $N$ discretized points, the truncated $u^*$ is $u^*_N=\sum_{k=-N/2}^{N/2-1}\hat{u}^*(k)e^{ikx}$, suppose 
\begin{equation}
\begin{gathered}
   \Tilde{\theta}^*_N \triangleq \arg\min_\theta \sum_{i=1}^{N} \Tilde{\mathcal{L}}_r\left[\hat{u}^\theta(k_i)\right].
\end{gathered}
\end{equation}
Then 
\begin{equation}
\label{eq: sp}
        \|u^{\Tilde{\theta}^*_N}-u^*\|_\Omega \leq \|u^{\Tilde{\theta}^*_N}- u^{*}_N\|_\Omega+\underset{\text{spectral error}}{\| u^{*}_N-u^*\|_\Omega} 
\end{equation}
As the capability of MLP is powerful enough, $\|u^{\Tilde{\theta}^*_N}- u^{*}_N\|_\Omega \leq \sum_{k=-N/2}^{N/2-1}\|\hat{u}^{\Tilde{\theta}^*_N}(k)-\hat{u}^{*}(k)\|=0$. Furthermore, as the spectral error is exponential convergence~\cite{Canuto1988Spectral}, then:
\begin{equation}
     \|u^{\Tilde{\theta}^*_N}-u^*\|_\Omega = o(N^{-s}), ~~\forall s>0.
\end{equation}

Thus, the convergence rate of SINNs is $o(N^{-s})$ for any $s>0$ while the convergence rate of PINNs is $\mathcal{O}(N^{-1/2})$. 

\section{Experiments}\label{sec:Experiment}
To demonstrate the performance of the proposed SINNs, we consider training the linear equations (\cref{table:linear}) using the framework shown in \cref{section: linear} and training the non-linear equations (\cref{table:nonlinear}) using the framework shown in \cref{section: IncomFlows}.
The details of the equations and training are available in \cref{Appendix: details}, the details of the relative L2 error metric used in our experiments are described in \cref{Relative error}. 
\begin{table}[ht]
\caption{Comparison of the relative errors for the linear equations}
\label{table:linear}
\centering
\scalebox{0.7}{
\begin{tabular}{clllll}
\toprule
 & equation                    &   & PINN & SINN (SI) &SINN (WL) \\\cmidrule(r){1-6}
\multirow{2}{*}{1-D}           & {\small convection-diffusion}        & $u$ & $1.79\times 10^{-2}\pm4.19\times 10^{-3}$    &  \boldmath{$1.14\times 10^{-5}\pm1.28\times 10^{-5}$}  &  $1.35\times 10^{-4}\pm1.64\times 10^{-4}$    \\
\cmidrule(r){3-6}
                              & diffusion                   & $u$ & $1.88\times 10^{-2}\pm1.44\times 10^{-2}$    &  $4.32\times 10^{-4}\pm1.61\times 10^{-4}$  &  \boldmath{$4.30\times 10^{-4}\pm2.38\times 10^{-4}$}    \\
                              \cmidrule(r){1-6}
\multirow{2}{*}{2-D}           & heat                        & $u$ & $1.06\times 10^{-3}\pm1.55\times 10^{-4}$    &  $1.09\times 10^{-3}\pm5.10\times 10^{-4}$  &  \boldmath{$3.21\times 10^{-4}\pm5.62\times 10^{-5}$}    \\
\cmidrule(r){3-6}
                              & heat\_random                & $u$ & $4.24\times 10^{-3}\pm4.93\times 10^{-3}$    &  \boldmath{$4.21\times 10^{-3}\pm9.79\times 10^{-4}$}  &  $1.97\times 10^{-2}\pm6.20\times 10^{-3}$    \\
                              \cmidrule(r){1-6}
\multirow{1}{*}{3-D}           & heat                    & $u$ & $1.19\times 10^{-1}\pm3.61\times 10^{-3}$    &  $7.29\times 10^{-2}\pm1.98\times 10^{-4}$  &  \boldmath{$4.70\times 10^{-2}\pm5.44\times 10^{-3}$}    \\
\bottomrule
\end{tabular}}
\end{table}
\begin{table}[ht]
\caption{Comparison of the relative errors for the non-linear equations}
\label{table:nonlinear}
\centering
\begin{tabular}{cllll}
\toprule
 & equation                    &   & PINN & SINN (SI) \\ \cmidrule(r){1-5}
\multirow{1}{*}{1-D}           & Burgers                     & $u$ & $5.17\times 10^{-4}\pm1.50\times 10^{-4}$    &  \boldmath{$1.62\times 10^{-4}\pm4.70\times 10^{-5}$}    \\
\cmidrule(r){1-5}
\multirow{4}{*}{2-D}           & \multirow{2}{*}{NS\_TG}     & $u$ & $6.97\times 10^{-4}\pm2.61\times 10^{-5}$    &  \boldmath{$5.63\times 10^{-4}\pm1.99\times 10^{-4}$}   \\
                              &                             & $v$ & $6.74\times 10^{-4}\pm3.26\times 10^{-4}$   &  \boldmath{$5.13\times 10^{-4}\pm1.71\times 10^{-4}$ }  \\
                              \cmidrule(r){3-5}
                              & \multirow{2}{*}{NS\_random} & $u$ & $2.20\times 10^{-3}\pm7.78\times 10^{-4}$    &  \boldmath{$1.30\times 10^{-3}\pm6.66\times 10^{-5}$}     \\
                              &                             & $v$ & $2.06\times 10^{-2}\pm8.58\times 10^{-3}$    &  \boldmath{$1.53\times 10^{-2}\pm6.82\times 10^{-4}$}    \\
\bottomrule
\end{tabular}
\end{table}

\subsection{Different order of derivatives}\label{sec:highorder}
To compare the efficiency between our SINNs and PINNs, we consider a specific one-dimensional (1-D) hyper-diffusion equation with different order of derivatives: 
\begin{equation}
\begin{gathered}\label{eq: highorder}
    \frac{\partial u}{\partial t}-\epsilon\frac{\partial^p u}{\partial x^p}=0, 
    \quad x\in \left[0,2\pi\right], t \in \left[0,T\right], \\
    u(0,x)=\sum_{k=0}^{N-1}\sin(kx),
\end{gathered}
\end{equation}
where $p$ is the order of the spacial derivatives. To balance the solution of different order, we set $\epsilon= 0.2^p, T=0.1$ in our experiments. The results are shown in \cref{fig:memory} and \cref{table: differenorder}.
\begin{table}[ht]
\caption{Comparison of the experiments for \cref{eq: highorder} with different order. '-' means we can't obtain the value because of the limitations of training time.}
\label{table: differenorder}
\centering
\scalebox{0.95}{
\begin{tabular}{cllll}
\toprule
\multicolumn{1}{l}{}    & $p$  & Relative error & Training time (hours)                        & Memory allocated (GB) \\
\midrule
\multirow{5}{*}{PINN}   & 2  & $7.35\times 10^{-2}\pm7.07\times 10^{-4}$         & 0.17                                    & 1.18             \\
                        & 4  & $1.10\times 10^{-2}\pm7.30\times 10^{-3}$         & 0.5                                     & 2.81             \\
                        & 6  & $2.16\times 10^{-2}\pm 1.75\times 10^{-3}$          & 4.39                                    & 9.06             \\
                        & 8  &     -           & >30000                     & 26.65            \\
                        & 10 &      -          &         -                                & 76.71            \\\midrule
\multirow{5}{*}{SINN (SI)} & 2  & $2.06\times 10^{-3}\pm1.51\times 10^{-3}$         & 0.33                                    & 1.1948           \\
                        & 4  & $1.67\times 10^{-3}\pm3.25\times 10^{-4}$         & 0.33                                    & 1.1948           \\
                        & 6  & $1.23\times 10^{-2}\pm1.36\times 10^{-3}$           & 0.33                                    & 1.1948           \\
                        & 8  & $8.33\times 10^{-3}\pm1.91\times 10^{-4}$          & 0.33                                   & 1.1948           \\
                        & 10 & $5.40\times 10^{-3}\pm2.33\times 10^{-4}$           & 0.35                                    & 1.1948       \\  
\midrule
\multirow{5}{*}{SINN (WL)} & 2  & $8.47\times 10^{-4}\pm3.29\times 10^{-4}$             & 0.36                                    & 1.1968           \\
                        & 4  & $1.25\times 10^{-2}\pm7.79\times 10^{-3}$           & 0.36                                    & 1.1968           \\
                        & 6  & $1.41\times 10^{-2}\pm1.30\times 10^{-3}$             & 0.35                                    & 1.1968           \\
                        & 8  & $8.39\times 10^{-3}\pm2.87\times 10^{-4}$            & 0.36                                    & 1.1968           \\
                        & 10 & $5.48\times 10^{-3}\pm1.55\times 10^{-4}$             & 0.36                                    & 1.1968       \\  
\bottomrule
\end{tabular}}
\end{table}

\paragraph{Training time} One may argue that in \cref{table: differenorder}, for the most general derivative term $p=2$, PINNs are more efficient than SINNs. It is because SINNs have the imaginary part thus the output channels are double the output channels of PINNs. However, we have a more comprehensive comparison in \cref{Appendix: time}, and the conclusion is: if the spacial derivative terms are more than one second-order derivative, including one third-order derivative, or one second-order derivative plus one first-order derivative, our SINNs are more efficient than PINNs; otherwise, PINNs are more efficient.

\subsection{Different spectral structures}
To verify the approximation of SINNs is not only for the low-frequency solutions, experiments on the diffusion equation (\cref{eq: different structure}) are implemented with different $N$:
\begin{equation}\label{eq: different structure}
\begin{gathered}
    u_t+au_x-\epsilon u_{xx}=0, x\in \left[0,2\pi\right], t \in \left[0,T\right],\\
    u(0,x)=\sum_{k=0}^{N-1} \sin\left(kx\right).
\end{gathered}
\end{equation}

\begin{figure}[!htbp]
  \centering
  \subfigure[\label{N_test_1} Relative error for different $N$]{\includegraphics[width=0.329\linewidth]{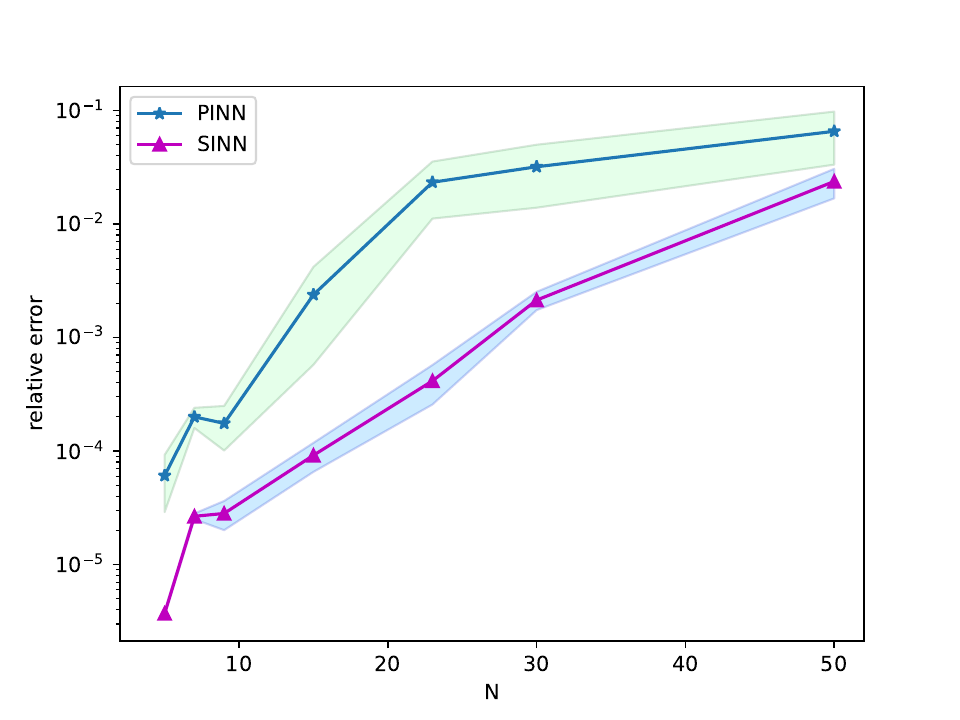}}
  \subfigure[\label{N_test_2} $u(0.1,x)$ for $N=23$]{\includegraphics[width=0.329\linewidth]{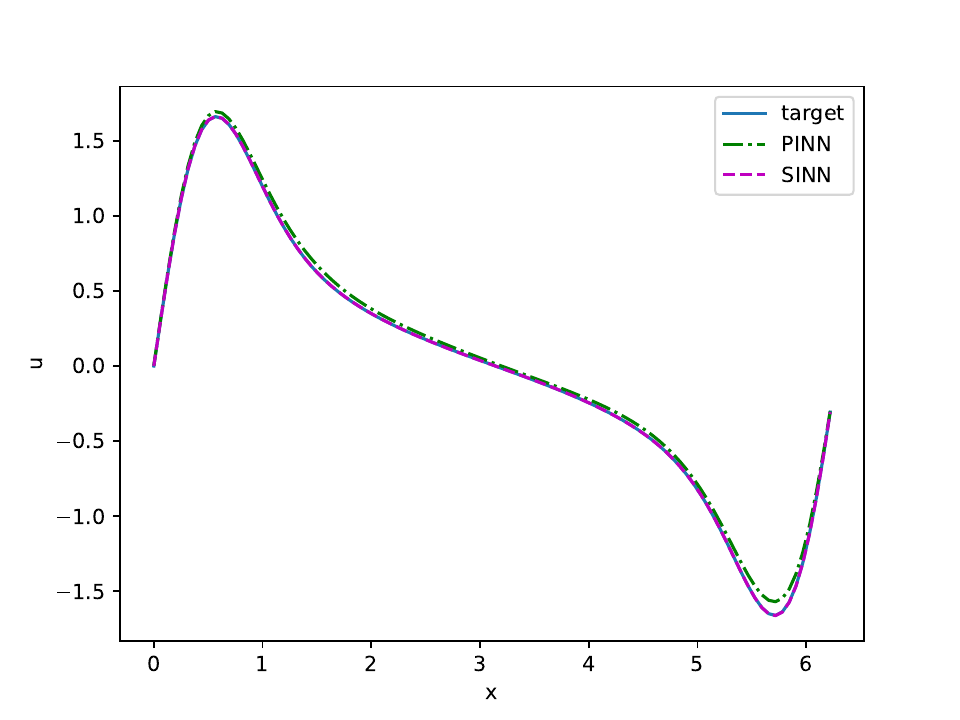}}
  \subfigure[\label{N_test_3} $u(0.1,x)$ for $N=30$]{\includegraphics[width=0.329\textwidth]{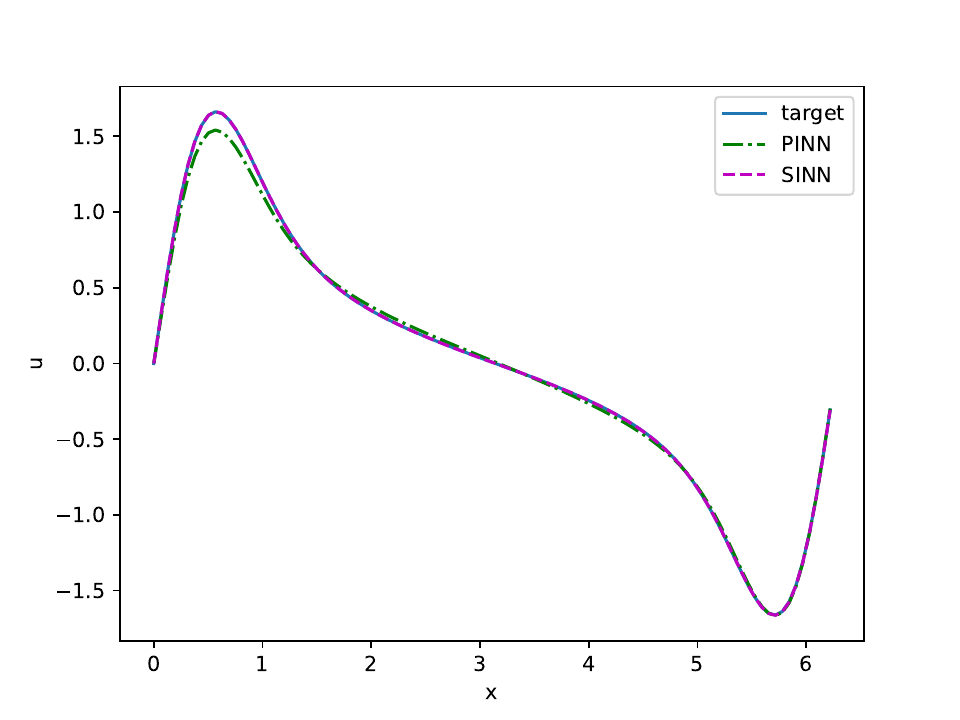}} 
  \caption{In experiments for the diffusion equation \cref{eq: different structure} with different $N$ in $u(0.1,x)$, the relative error with $N$ is plotted in \subref{N_test_1}.
  And the solutions of $u(0.1,x)$ for $N=23$ and $30$ are shown in \subref{N_test_2} and \subref{N_test_3}, respectively.
  Those results indicate that SINNs are robust even with complex spectral structure.}
  \label{fig:N_testing}
\end{figure}	
Compared to PINNs,  \cref{fig:N_testing} indicates SINNs are more robust even with a more complex structure in spectral domain. More details including the results of WL for the diffusion equation with different $N$ are presented in \cref{table: diffusion}.

\subsection{Ablation study of the SI and WL}
SI and WL are different approaches designed for the same object, in this section, we provide an ablation study of how to choose the strategies of SI and WL, and how to set the corresponding hyperparameters. To tune the hyperparameters, one should consider the distribution of the Fourier coefficients: the coefficients concentrate more on the low-frequency, and you should have larger $\epsilon$ for WL or smaller $\alpha$ and smaller $\beta$. To select the approach, ideally, WL is better because this approach can learn all coefficients fully. However, as the relationship between the capability of networks and the complexity of PDEs is unknown, using WL requires additional consideration of the size of the network. If some low-frequencies are out of the capability, WL makes the network continue focusing on them because the hard constraint on the loss function makes the rest frequencies be multiplied by a small weight. So in practice, SI performs better thanks to its soft constraint on the structure of SINNs. We did a series experiments on diffusion equation to show the influence of hyperparameters. The results are depicted in ~\cref{wl_ablation,si_ablation} and the details can be found in \cref{Appendix: ablation}.

Additionally, although we think SI and WL have the same effect, thus using both of them isn't cost-effective, we did extra experiments that combine SI and WL strategies together to verify how SI and WL influence each other. We plot a snapshot of $\epsilon=10^{-5}$ in \cref{si_wl_comb} which including the optimal hyperparameters. The experiments shows that combining SI and WL is not an efficient way and sometimes slightly damages the accuracy.

\begin{figure}[!htbp]
  \centering
  \subfigure[\label{wl_ablation} Ablation of WL]{\includegraphics[width=0.329\linewidth]{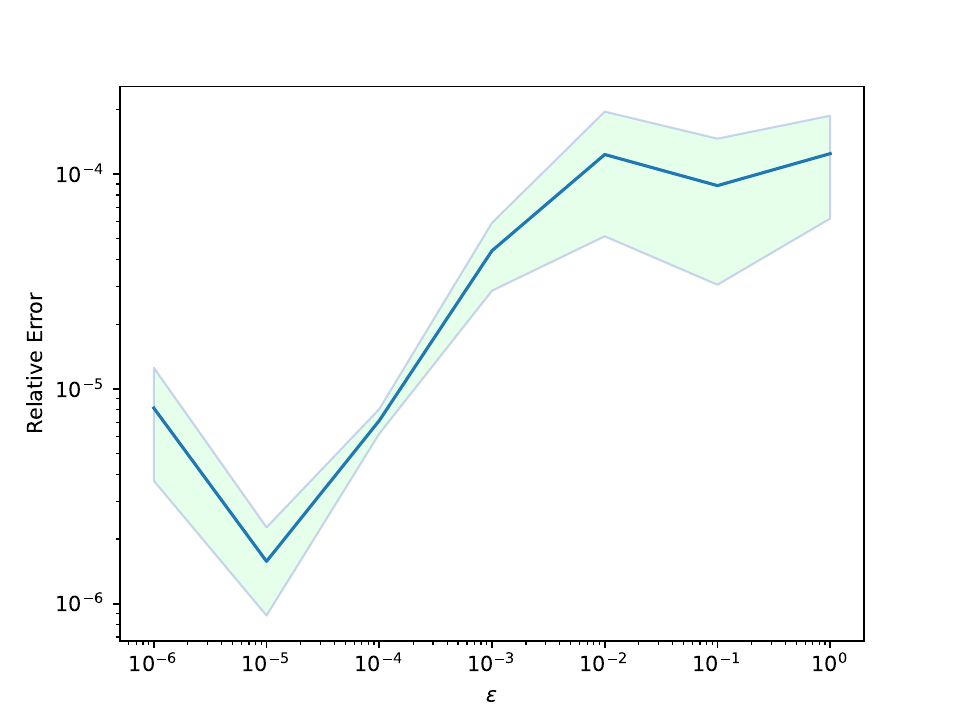}} 
  \subfigure[\label{si_ablation} Ablation of SI]{\includegraphics[width=0.329\linewidth]{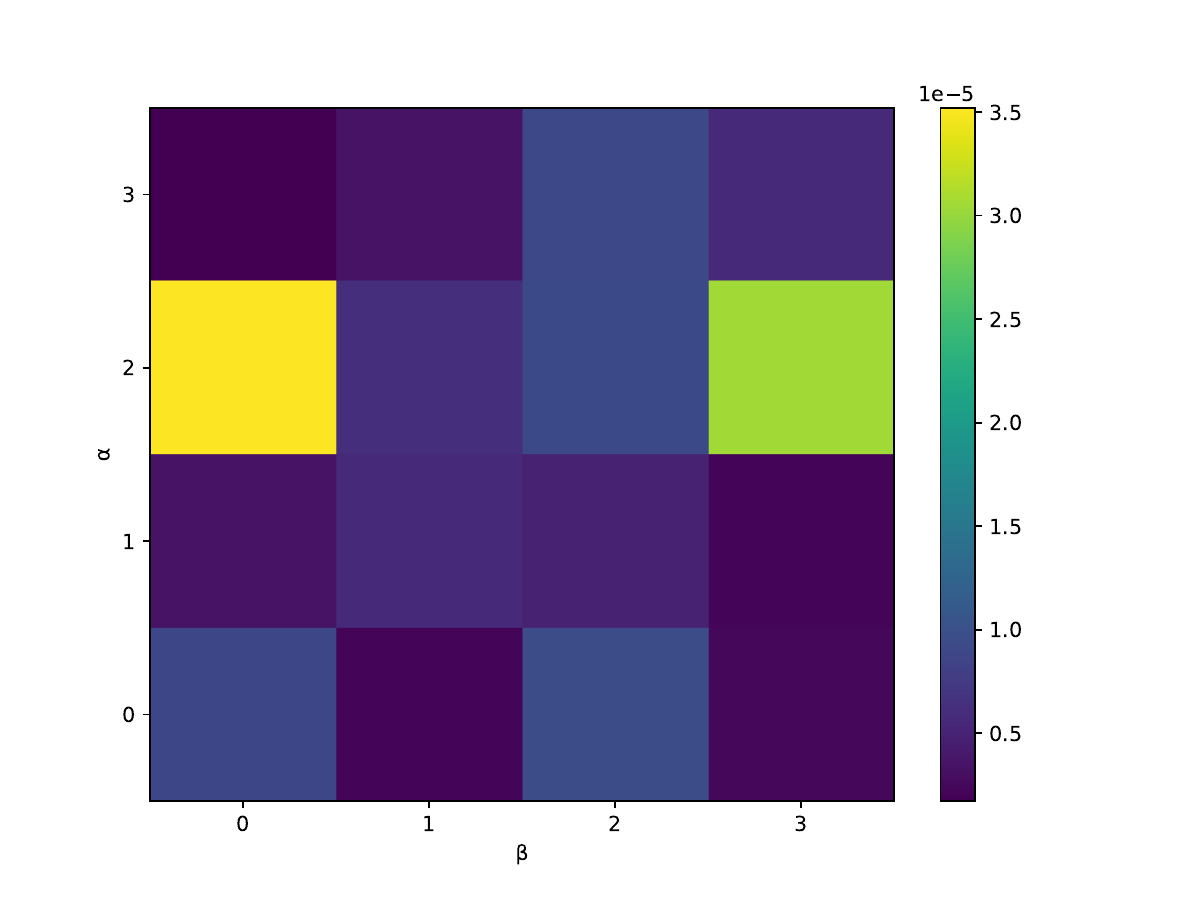}}
\subfigure[\label{si_wl_comb} Combination of WL and SI in $\epsilon=10^{-5}$]{\includegraphics[width=0.329\linewidth]{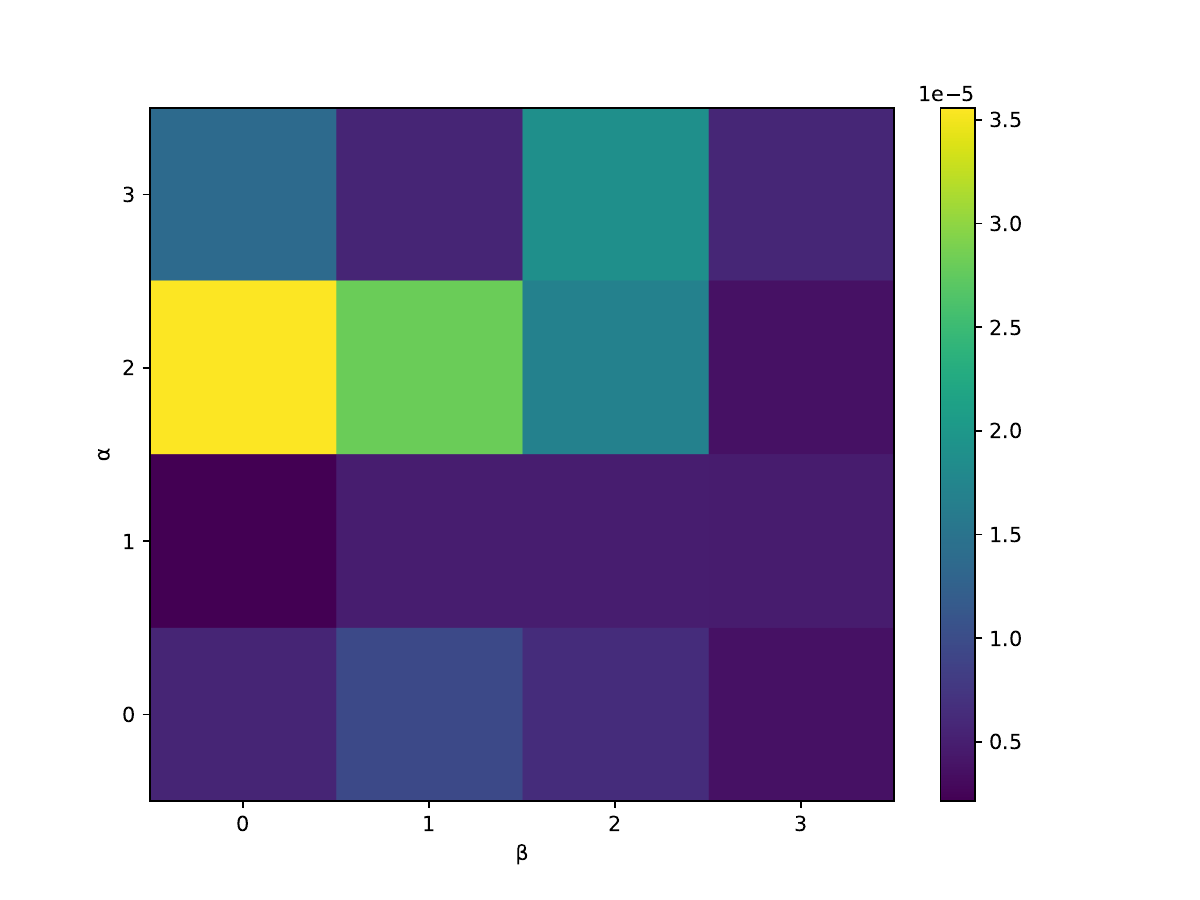}}
  \caption{Ablation study in diffusion equation. In \cref{wl_ablation}, the best hyperparameter is $\epsilon=10^{-5}$, and the corresponding error is $1.58\times 10^{-6}\pm6.94\times 10^{-7}$; in \cref{si_ablation}, the best hyperparameter for SI is $\alpha=3, \beta=0$, and the corresponding error is $1.73\times 10^{-6}\pm1.41\times 10^{-6}$; in \cref{si_wl_comb}, the best hyperparameter is $\alpha=1, \beta=0, \epsilon=10^{-5}$, and the corresponding error is $2.14\times 10^{-6}\pm\times 10^{-7}$.}
  \label{fig:ablation}
\end{figure}

\section{Conclusion and future work}\label{section: conlclusion}
In this study, the Spectral-Informed Neural Networks (SINNs) are developed to approximate the function in the spectral domain. We point out that the difference in learning between the physical domain and the spectral domain is the importance of input points, and therefore specific training strategies are introduced. The chosen Fourier basis helps us compute the spatial derivatives as well as train the neural network efficiently and with low memory allocation. Besides, SINNs have more accurate predictions due to the exponential accuracy of the spectral method. To provide evidence that SINNs have not only a notable reduction in training time but also significant improvements in accuracy, we did a series of experiments on linear and non-linear PDEs.
\paragraph{Limitations} The current SINNs also inherit the disadvantages of spectral methods, for some 
PDEs with complex geometries or detailed boundaries in more than one space variable would cause spectral methods to collapse, and so would SINNs. On the other hand, functions are often more complex in the spectral domain than in the physical domain, thus those functions that cannot be approximated by PINNs always cannot be approximated by SINNs with the same architecture. As for non-linear equations, in our experiments, SINNs are usually unstable when dealing with the aliasing error, due to the nonnegligible bias between \cref{SINN-NSEq} and the ideal loss that the aliasing error is zero. 
\paragraph{Future} Apart from the positive results shown in this paper, the above limitations remain to be investigated further in the future.  For the inherited disadvantages from spectral methods, in essence, the spectral method is a specific type of collocation methods that rely on selected basis functions satisfying boundary conditions. Similar to the spectral methods, the collocation methods ensure the residual of the target equation approaches zero at the collocation points associated with the basis functions. Therefore, the SINNs could be developed based on the valuable insights of earlier studies~\cite{Canuto1988Spectral} on collocation methods. To improve the learning capability of SINNs, we can replace the MLP with the architecture from Neural Operators, such as PINO~\cite{li2021physics} and physics-informed DeepOnets~\cite{wang2021learning}. With the direct derivatives on the spacial, approximating in the spectral domain can make Neural Operators more flexible in dealing with the physics-informed loss. As for the nonnegligible bias, with the bias decreases with the number of grids increases, fine enough grids can make the bias negligible.  

Furthermore, during conducting our experiments, we found that SINNs are more sensitive to the initial conditions, so assigning weights~\cite{wang2022and, MCCLENNY2023111722,anagnostopoulos2024learning} in $\Tilde{\mathcal{L}}_r$ and $\Tilde{\mathcal{L}}_{ic}$ may have a dramatic improvement on SINNs. Some tricks in classical spectral methods can be investigated, for example 1) we can investigate implementing Compressive Sampling~\cite{candes2006compressive, bayindir2016compressive} on SINNs to further reduce the training time; 2) we can investigate the smoothed series $S_N[u](t, x)=\sum_{k=-N / 2}^{N / 2-1} \sigma_k \hat{u}(t, k) e^{i k x}$ to handle PDEs with discontinuous solutions or sharp transitions.

\bibliography{ref}
\appendix
\section{Faà di Bruno's formula}\label{Appendix: FB_formula}
\begin{figure}[ht]
    \centering
    \includegraphics[width=0.45\linewidth]{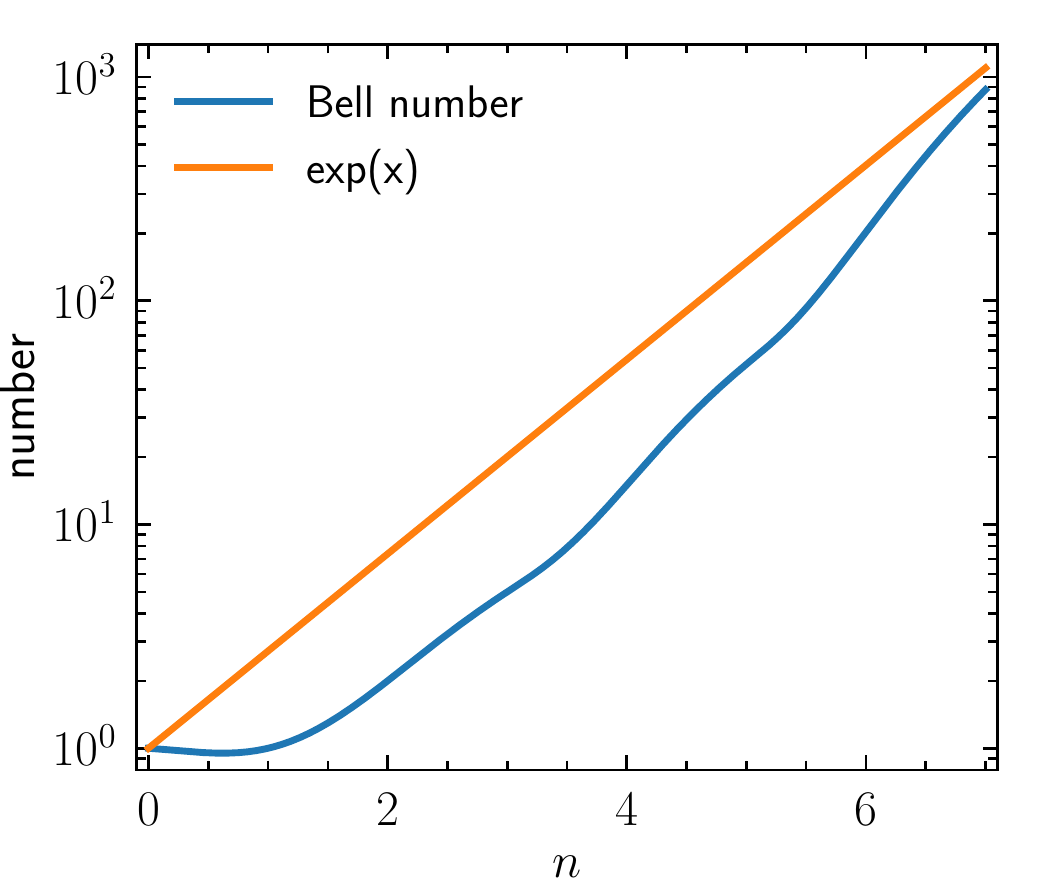}
    \caption{Comparison of the total number between the Bell number and the exponential function.}\label{fig:belln}
\end{figure}

Given $f,g$ are sufficiently smooth functions, Faà di Bruno's formula states that the $n$-th derivative of $f\left(g\left(x\right)\right)$ is a sum of products of various orders of derivatives of the component functions:
$$
\frac{d^n}{d x^n} f(g(x))=(f \circ g)^{(n)}(x)=\sum_{\pi \in \Pi} f^{(|\pi|)}(g(x)) \cdot \prod_{B \in \pi} g^{(|B|)}(x),
$$
where $\Pi$ is the set of all partitions of the set $\{1, \ldots, n\},|\pi|$ is the number of blocks in the partition $\pi$, and $|B|$ is the number of elements in the block $B$.
Since Faà di Bruno's formula is a generalization of the chain rule used in first-order, one can directly apply this formula to achieve accurate higher-order AD efficiently by avoiding a lot of redundant computations. However, the total number of partitions of an n-element set is the Bell number~\cite{lucas1990introduction}: $\#\{\Pi\}=B_n=\sum_{k=0}^n {\binom{n}{k}} B_k$ that increases exponentially with the order $n$ (\cref{fig:belln}). 
\section{Aliasing error}\label{Appendix: aliasing}
The quadratic non-linear term $w(x) = u(x)v(x)$ can be regarded as the convolution sum of the Fourier expansions of the corresponding two terms in spectral methods, \textit{i.e.},
\begin{equation}
  \hat{w}(k) = \sum\limits_{m+n=k}\hat{u}(m)\hat{v}(n) ,
\end{equation}
where $\hat{w}$, $\hat{u}$, and $\hat{v}$ represent the Fourier coefficients of $u(x)$, $v(x)$, and $w(x)$, respectively. In practical computations, when dealing with truncated Fourier series consisting of only $N$ components, a challenge arises: some frequencies of the summation of the two terms exceed the range of the truncated frequency. This issue stems from the periodic nature of the discrete Fourier transform, where components outside the frequency range are mapped back into the range as lower or higher frequency components, leading to aliasing errors in the nonlinear terms. For instance, components whose frequencies satisfy $m+n>N/2-1$ are erroneously identified as components of frequency $(m+n-N)$. Two truncation methods, the $2/3$ rule~\cite{Orszag1971dealias} and the Fourier smoothing method~\cite{Hou2007}, are designed to retain the components within the range of $-N/3$ to $N/3-1$ for the nonlinear terms. The $2/3$ rule sets the last $1/3$ of the high-frequency modes to zero, leaving the rest unchanged, while the Fourier smoothing method uses an exponential function $y=e^{-36 (|\boldsymbol{k}| / N)^{36}}$ to filter out the high-frequency modes. The Fourier smoothing method is employed in our SINNs since it can capture more effective Fourier modes and make a more accurate solution~\cite{Hou2007}.

\section{Details of Experiments}\label{Appendix: details}
In the following experiments, we proceed by training the model via stochastic gradient descent using the Adam~\cite{kingma2014adam} optimizer with the exponential decay learning rate. The hyperparameters for exponential decay are: the initial learning rate is $10^{-3}$, the decay rate is $0.95$ and the number of transition steps is 10000. The MLP is equipped with the Sigmoid Linear Unit (SiLU) activations and Xavier initialization.

Note that there is no injection of external source terms in our experiments, resulting in a decay of the quantities, including temperature and energy, over time. As time increases, the related functions gradually become smoother, and the overall flow field tends to be constant. Herein, to clearly demonstrate the advantages and performance of our SINNs, the temporal domain is restricted to the interval when the flow field undergoes significant changes. Additionally, we did an experiment in a long temporal domain for the 1-D convection-diffusion equation with periodic boundary conditions.

\subsection{The experiments on linear equations}
\subsubsection{1-D problems}
In 1-D problems, we discretize the spatial domain to 100 points and the temporal domain to 100 points, thus the total size of the discretization for PINNs is $100 \times 100$. Thanks to the symmetric of real functions in the spectral domain, the total size of the discretization for SINNs is $51 \times 100$. The MLP we used for both PINNs and SINNs is $10 \times 100$: 10 layers and every hidden layer has 100 neurons. We train both PINNs and SINNs for $5\times10^5$ iterations.
\paragraph{Convection-diffusion equation (convection-diffusion)}
Our first experiment is the 1-D convection-diffusion equation with periodic boundary conditions, and the convection-diffusion equation can be expressed as follows:
\begin{equation}
\begin{gathered}
    u_t+au_x-\epsilon u_{xx}=0, x\in \left[0,2\pi\right], t \in \left[0,T\right],\\
    u(0,x)=\sum_{k=0}^{N-1} \sin\left(kx\right),
\end{gathered}
\end{equation}
with the analytic solution 
\begin{equation}
    u(t,x)=\sum_{k=0}^{N-1} \sin\left(kx - kat\right)  e^{-\epsilon  k^2 t},
\end{equation}
  where $T=0.1$, $\epsilon=0.01$, $a=0.1$, and $N=6$ in our experiments.

Additionally, to verify our methods on long temporal domain, we did two experiments one is $T=1$ (discretize to 100 points) and another is $T=10$ (discretize to 1000 points).

\begin{table}[ht]
\caption{Long temporal domain of $T=1$}
\label{table:long domian t=1}
\centering
\scalebox{0.9}{
\begin{tabular}{cllllll}
\toprule
$t$(s) & 0.10 & 0.30 & 0.50 & 0.70 & 0.90 & 1.00 \\\cmidrule(r){1-7}
relative error   &$1.43\times 10^{-5}$ &$1.55\times 10^{-5}$ &$ 1.39\times 10^{-5}$ &$ 1.44\times 10^{-5}$ &$ 1.47\times 10^{-5}$ &$ 1.49\times 10^{-5}$
\\
\bottomrule
\end{tabular}}
\end{table}

\begin{small}
\begin{table}[ht]
\caption{Long temporal domain of $T=10$}
\label{table:long domian t=10}
\centering
\scalebox{0.9}{
\begin{tabular}{cllllll}
\toprule
$t$(s) & 1.00 & 3.00 & 5.00 & 7.00 & 9.00 & 10.00 \\\cmidrule(r){1-7}
relative error    &$8.86\times 10^{-5}$ &$6.48\times 10^{-5}$ &$ 6.59\times 10^{-5}$ &$ 4.98\times 10^{-5}$ &$ 3.96\times 10^{-5}$ &$ 4.21\times 10^{-5}$
\\
\bottomrule
\end{tabular}}
\end{table}
\end{small}
\begin{figure}[!htbp]
    \centering
    \includegraphics[width=1\linewidth]{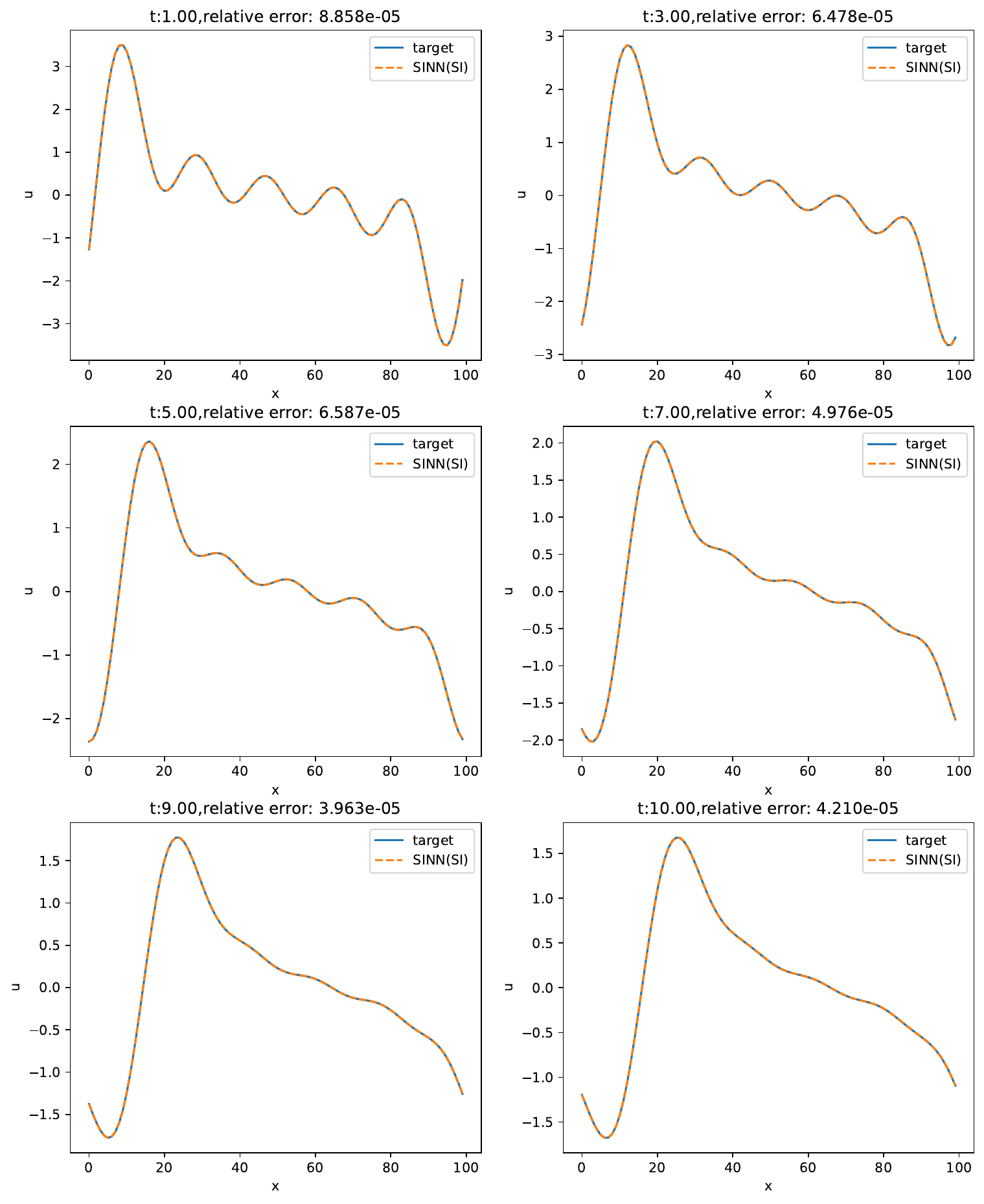}
    \caption{Lone temporal domain experiment of $T=10$, with the time $t$ increases, the function becomes smoother and the relative error becomes even smaller. Furthermore, the error is consistency on time showing there is no error accumulation on SINNs }
    \label{fig:long}
\end{figure}
\paragraph{Diffusion equation (diffusion)}
Another set of experiments on 1-D linear equations is about the diffusion equation, which can be written as
\begin{equation}
\begin{gathered} \label{eq:diffusion}
    u_t=\epsilon u_{xx}, \quad x\in \left[0,2\pi\right], ~~t \in \left[0,T\right],\\
    u(0,x)=\sum_{k=0}^{N-1}\sin(kx),
\end{gathered}
\end{equation}
  with the analytic solution 
\begin{equation}
    u(t,x)=\sum_{k=0}^{N-1}\sin(kx)e^{-\epsilon k^2t} , \label{eq: sol_diffusion}
\end{equation}
  where $T=0.1$ and $\epsilon=1.0$ in our experiments.
Based on PINNs and SINNs with sampling by importance, \cref{fig:N_testing} illustrates two groups of experiments with varied $N$ 
Besides, \cref{table:linear} shows the experimental results with $N=23$, while \cref{table: diffusion} presents more results with different $N$.

\begin{table}[ht]
\caption{Comparison of the relative errors for the diffusion equation with different N in \cref{eq: sol_diffusion}}
\label{table: diffusion}
\centering
\scalebox{0.9}{
\begin{tabular}{clll}
\toprule
  $N$  & PINN & SINN (SI) & SINN (WL) \\
  \midrule
5   & $6.07\times 10^{-5}\pm3.17\times 10^{5}$    &  \boldmath {$3.71\times 10^{-6}\pm1.59\times 10^{-7} $}  &    $7.94\times 10^{-5}\pm1.80\times 10^{-5}$    \\\midrule
7   & $2.00\times 10^{-4}\pm4.04\times 10^{5}$    &  \boldmath {$2.66\times 10^{-5}\pm1.54\times 10^{-6} $}  &    $1.46\times 10^{-4}\pm5.36\times 10^{-5} $    \\\midrule
9   & $1.75\times 10^{-4}\pm7.42\times 10^{5}$    &  \boldmath {$2.82\times 10^{-5}\pm8.11\times 10^{-6} $}  &    $6.97\times 10^{-5}\pm9.18\times 10^{-6} $    \\\midrule
15  & $2.38\times 10^{-3}\pm1.80\times 10^{3}$   &  \boldmath {$9.17\times 10^{-5}\pm2.62\times 10^{-5} $}  &    $1.44\times 10^{-4}\pm4.27\times 10^{-5}$    \\\midrule
23  & $2.33\times 10^{-2}\pm1.21\times 10^{2}$    &             \boldmath {$4.15\times 10^{-4}\pm1.58\times 10^{-4}$}  &    $4.30\times 10^{-4}\pm2.38\times 10^{-4}$   \\\midrule
30  & $3.19\times 10^{-2}\pm1.80\times 10^{2}$    &  \boldmath {$2.13\times 10^{-3}\pm3.91\times 10^{-4} $}  &    $5.40\times 10^{-1}\pm3.00\times 10^{-1} $    \\\midrule
50  & $6.55\times 10^{-2}\pm3.22\times 10^{2}$    &  \boldmath {$2.37\times 10^{-2}\pm6.89\times 10^{-3} $}  &    $7.59\times 10^{-1}\pm2.28\times 10^{-1} $    \\
\bottomrule
\end{tabular}}
\end{table}

One may observe that the weighted loss method fails when $N>30$. Because the weighted loss forces SINNs to pay more attention on the low-frequency part, SINNs with weighted loss will abandon the high-frequency if the loss on low-frequency is not small enough. 
\subsubsection{2-D problems}
In 2-D problems, the MLP we used for both PINNs and SINNs is $10 \times 100$: 10 layers and every hidden layer has 100 neurons. We train both PINNs and SINNs for $10^6$ iterations.
\paragraph{Heat equation with analytic solution (heat\_analytic)}
For a 2-D linear problem, the heat equation with the following initial condition is considered here:
\begin{equation}
\begin{gathered}
        u_t=\epsilon\left(u_{xx}+u_{yy}\right), \quad \boldsymbol{x} \in [0,2\pi]^2, \ t \in [0,T], \\
        u(0,\boldsymbol{x})=\sum_{k=0}^{N-1} \left[\sin\left(kx\right) + \sin\left(ky\right)\right],
\end{gathered}
\end{equation}
with the analytic solution 
\begin{equation}
    u(t,x)=\sum_{k=0}^{N-1} \left[\sin\left(kx\right) + \sin\left(ky\right)\right]e^{- \epsilon k^2t} ,
\end{equation}
  where  $T=0.01$, $\epsilon=1.0$, and $N=10$ in our experiment.
The discretization of the spatial and temporal domains is set to $100 \times 100$ and $10$ points, respectively.
Thus, the total size of the discretization for PINNs is $100 \times 100 \times 10$, while the total size for SINNs can be reduced to $51 \times 100 \times 10$ due to the Fourier transform for real functions. 
\paragraph{Heat equation with random initialization (heat\_random)}
The 2-D heat equation with the Gaussian random initial condition is included here, which can be written as
\begin{equation}\label{eq: 2-D heat_random}
\begin{gathered}
        u_t=\epsilon\left(u_{xx}+u_{yy}\right), \quad \boldsymbol{x} \in [0,2\pi]^2, \ t \in [0,T], \\
        \hat{u}(0,\boldsymbol{k})=\hat{g}(\boldsymbol{k}),  \\
        \hat{g}(\boldsymbol{k}) = \left\{ \begin{aligned}
            10^4 \sqrt{ 0.123456 / H(1)} h(\boldsymbol{k}) & , ~~{\frac{1}{2}} \leq |\boldsymbol{k}| < {\frac{3}{2}} ,\\
            10^4 \sqrt{ 0.654321 / H(2)} h(\boldsymbol{k}) & , ~~{\frac{3}{2}} \leq |\boldsymbol{k}| < {\frac{5}{2}} ,\\
            10^4 \sqrt{ 0.345612 / H(3)} h(\boldsymbol{k}) & , ~~{\frac{5}{2}} \leq |\boldsymbol{k}| < {\frac{7}{2}} ,\\
            10^4 \sqrt{ 0.216543 / H(4)} h(\boldsymbol{k}) & , ~~{\frac{7}{2}} \leq |\boldsymbol{k}| < {\frac{9}{2}} ,\\
            10^4 \sqrt{ 0.561234 / H(5)} h(\boldsymbol{k}) & , ~~{\frac{9}{2}} \leq |\boldsymbol{k}| < {\frac{11}{2}} ,\\
            10^4 \sqrt{ 0.432165 / H(6)} h(\boldsymbol{k}) & , ~~{\frac{11}{2}} \leq |\boldsymbol{k}| < {\frac{13}{2}} ,\\
            0 & , ~~ |\boldsymbol{k}| \geq {\frac{13}{2}} ,
        \end{aligned} \right.\\
        H(n) = \sum\limits_{n-{\frac{1}{2}}\leq |\boldsymbol{k}|< n+{\frac{1}{2}}} |h(\boldsymbol{k})|^2 ,\\
\end{gathered}
\end{equation}
where $h \in \mathbb{C}$ generated by standard normal distribution fulfills the symmetry $h(\boldsymbol{k})=\bar{h}(-\boldsymbol{k})$, and the parameters $T=0.01$, $\epsilon=1.0$, and $N=10$ are taken in our experiment.
The spatial and temporal domains are discretized to $100 \times 100$ and $6$ points, respectively.
The total size of the discretization for PINNs is $100 \times 100 \times 6$, while the total size for SINNs is reduced to $51 \times 100 \times 6$ since the functions are real.
The solutions $u(0.01, \boldsymbol{x})$ in our experiments for the 2-D heat equation with the Gaussian random initial condition are plotted in \cref{fig:heat_random}.
The spectral method in \cref{Appendix: spectral method} computes the numerical solution $u(0.01, \boldsymbol{x})$ with a sufficiently small time step.

\begin{figure}[!htbp]
  \centering
  \subfigure[\label{heat_random_1} Exact]{\includegraphics[width=0.329\linewidth]{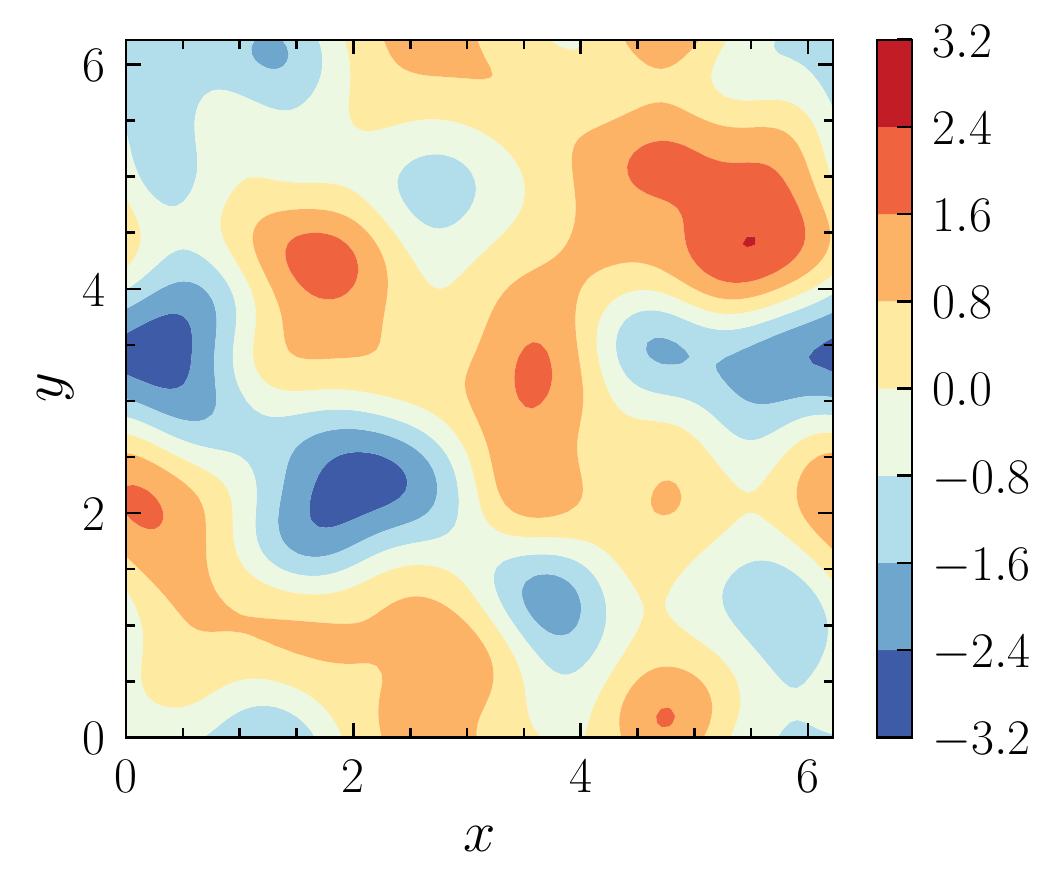}}
  \subfigure[\label{heat_random_2} PINNs]{\includegraphics[width=0.329\linewidth]{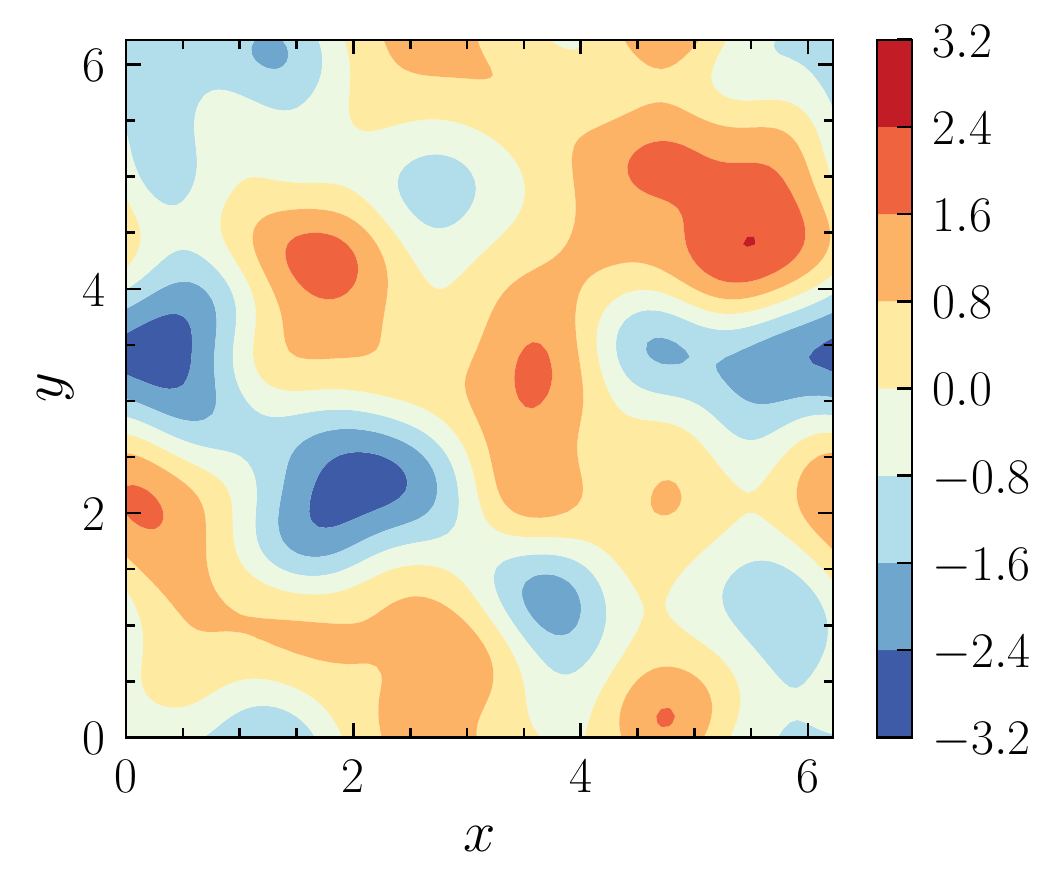}}
  \subfigure[\label{heat_random_3} SINNs]{\includegraphics[width=0.329\textwidth]{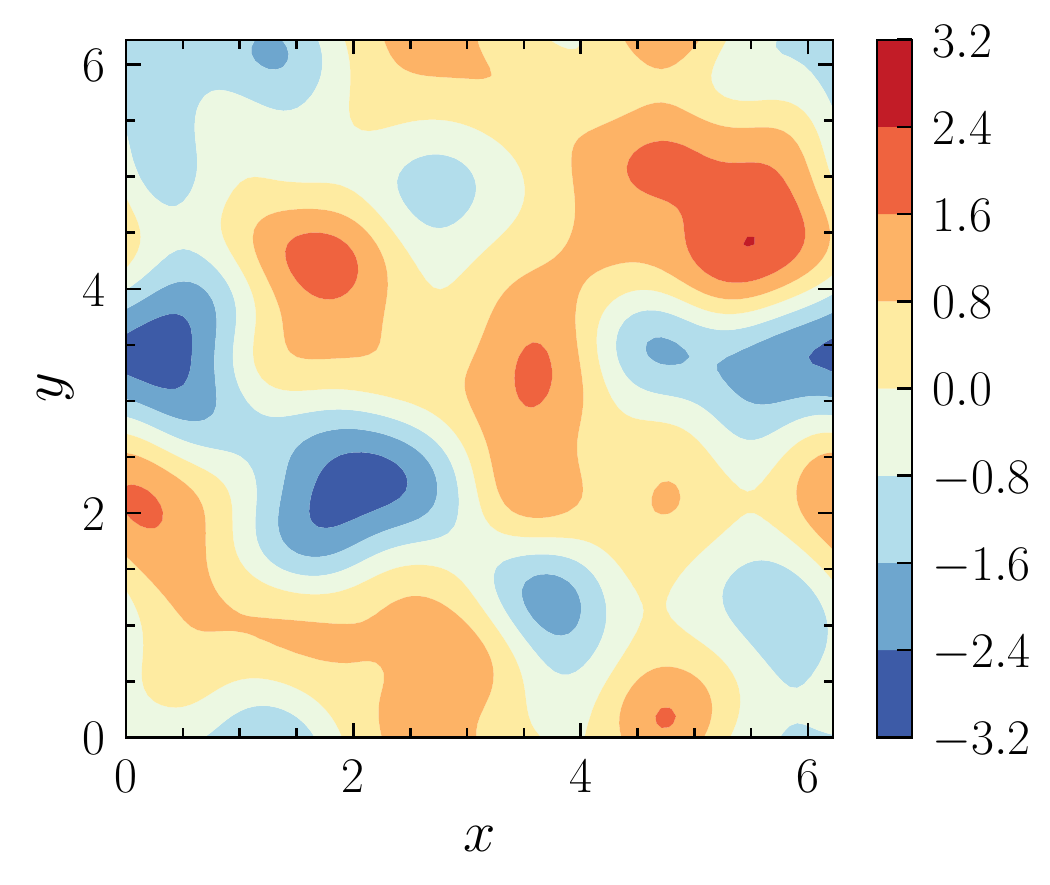}} 
  \caption{The exact solution, the predicted solution by PINNs, and the predicted solution by SINNs for $u(0.01, \boldsymbol{x})$ in the 2-D heat equation with the Gaussian random initial condition are displayed in \subref{heat_random_1}, \subref{heat_random_2}, and \subref{heat_random_3}, respectively.}
  \label{fig:heat_random}
\end{figure} 
\subsubsection{3-D problems}
\paragraph{Heat equation with analytic solution (heat\_analytic)}
To compare SINNs and PINNs for 3-D linear equations, a heat equation problem is considered here, which has the form
\begin{equation}
\begin{gathered}
    u_t=\epsilon\left(u_{xx}+u_{yy}+u_{zz}\right), \quad \boldsymbol{x} \in [0,2\pi]^3, \ t \in [0,T],\\
    u(0,\boldsymbol{x})=\sum_{k=0}^{N-1} \left[\sin\left(kx\right) + \sin\left(ky\right)+sin\left(kz\right)\right],
    \end{gathered}
\end{equation}
with the analytic solution 
\begin{equation}
    u(t,\boldsymbol{x})=\sum_{k=0}^{N-1}  \left[\sin\left(kx\right) + \sin\left(ky\right)+\sin\left(kz\right)\right]e^{- \epsilon k^3t}.
\end{equation}
  where $T=0.01$, $\epsilon=1.0$, and $N=5$ in our experiment.
The discretized spatial and temporal domains are $100 \times 100 \times 100$ and $10$ points, respectively.
The total size of the discretization for PINNs is $100 \times 100 \times 100 \times 10$, while the total size for SINNs is decreased to $51 \times 100 \times 100 \times 10$ for the real function $u$. 
\subsection{The experiments on non-linear equations}
\subsubsection{1-D problems}
\paragraph{Burgers equation (Burgers)}
One of the most important 1-D nonlinear equations is the Burgers equation, taking the following form:
\begin{equation}\label{eq: burgers}
\begin{gathered}
        u_t = \nu u_{xx} - u u_x, \quad x\in \left[0,2\pi\right], t \in \left[0,T\right],\\
        u(0,x)=\sum_{k=1}^N \sin (kx)
\end{gathered}
\end{equation}
  where $T=0.1, N=3 $ and $\nu= \pi / 150$ in our experiment.
The discretization of the spatial and temporal domains is set to $100$ and $11$ points, respectively.
The total size of the discretization for PINNs is $100 \times  11$, while the total size for SINNs is $51 \times 11$ for a real $u$. 

\subsubsection{2-D problems}
\paragraph{NS equations with Taylor–Green vortex (NS\_TG)}
The 2-D nonlinear NS equation is the same as \cref{eq:ns} with $T=2$, $\nu=2\pi/100$, and $\boldsymbol{\boldsymbol{g}(\boldsymbol{x})}=\left(-\cos(x)\sin(y),\ \sin(x)\cos(y)\right)$ in our experiment.
The spatial and temporal domains are discretized to $100 \times 100$ and $11$ points, respectively.
The total size for PINNs is $100 \times 100 \times 11$, while the total size for SINNs is $51 \times 100 \times 11$ since $\boldsymbol{u}$ is real.

\begin{figure}[!htbp]
  \centering
  \subfigure[\label{ns_random_1} Exact u]{\includegraphics[width=0.329\linewidth]{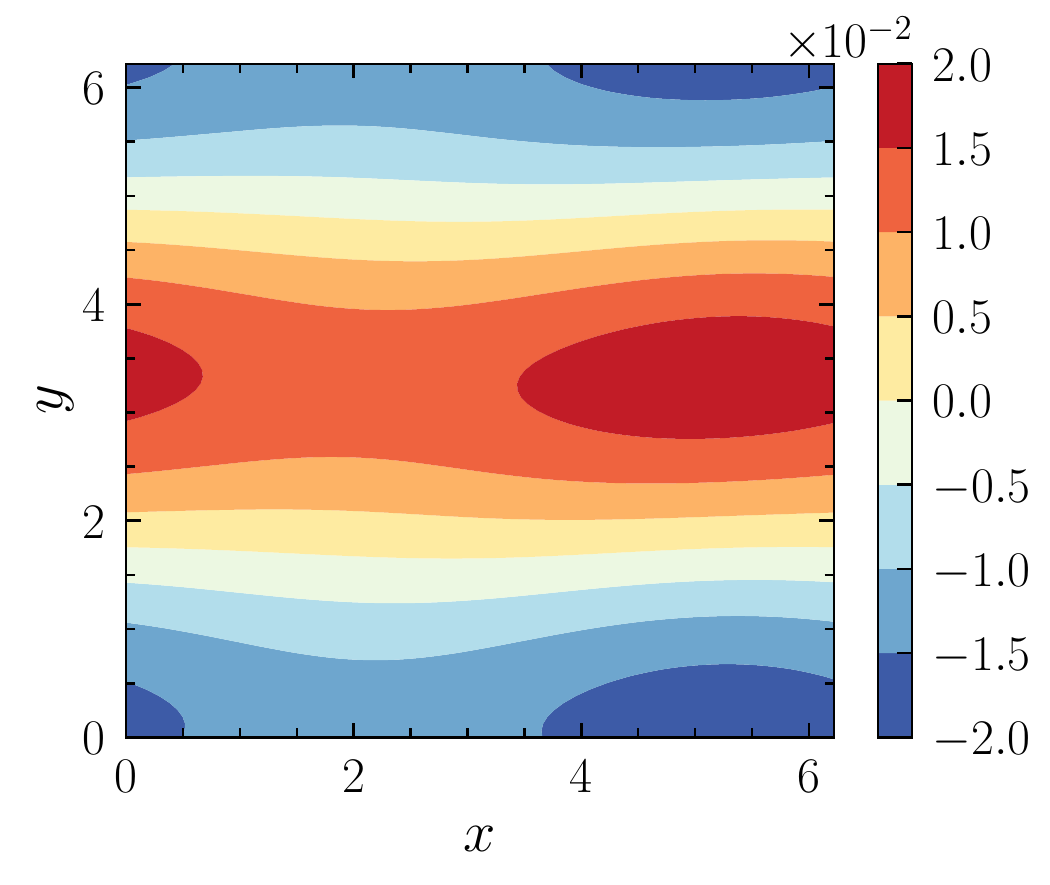}}
  \subfigure[\label{ns_random_2} Predicted u]{\includegraphics[width=0.329\linewidth]{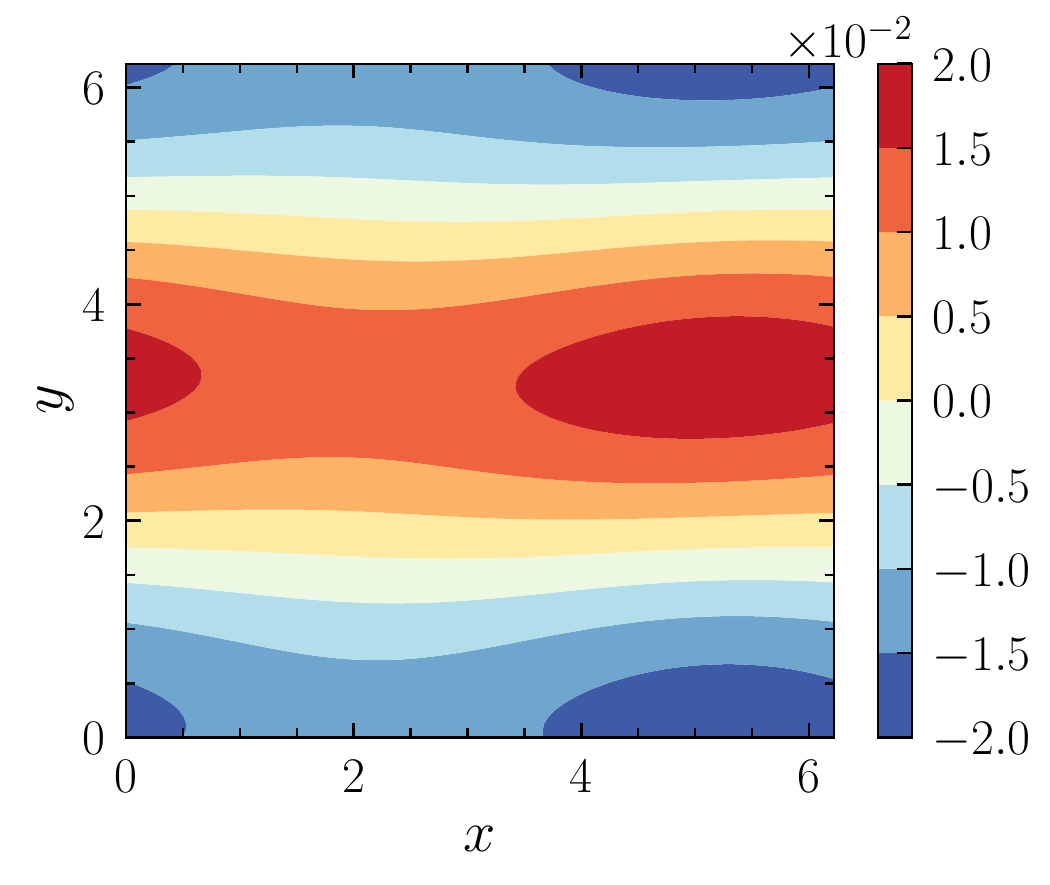}}
  \subfigure[\label{ns_random_3} Squared error of u]{\includegraphics[width=0.329\textwidth]{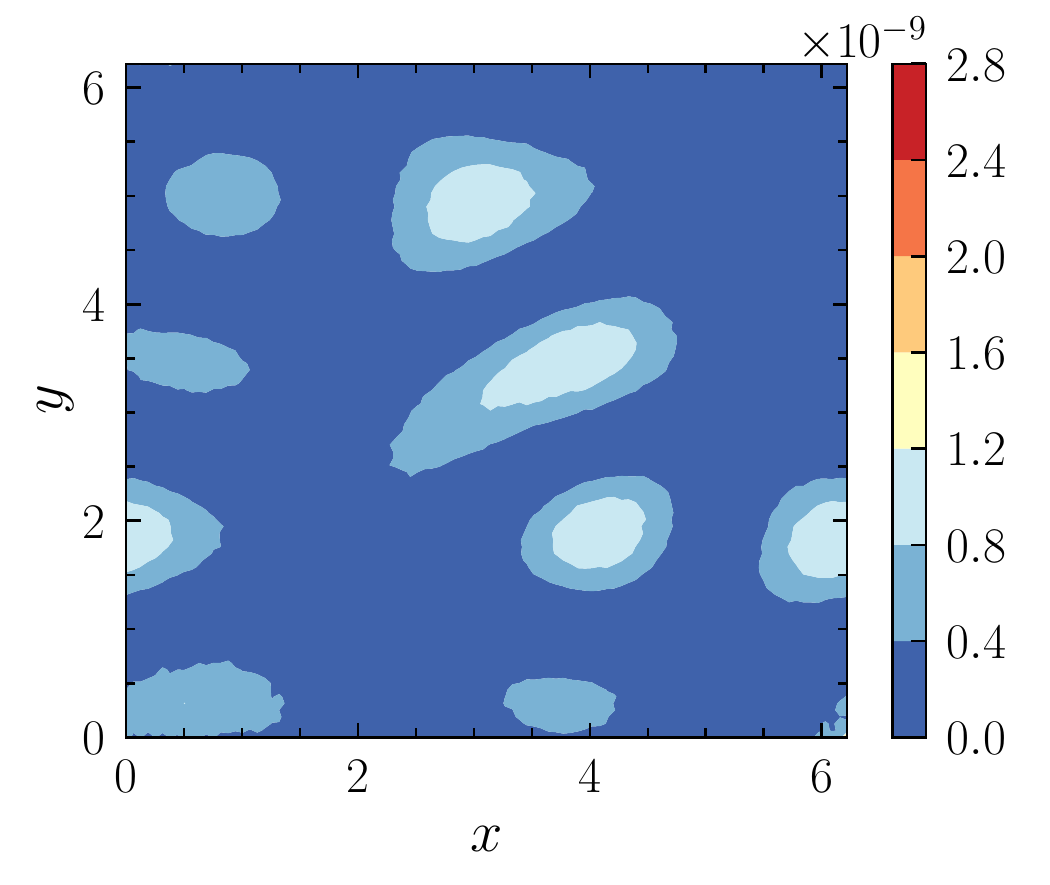}} 
     \subfigure[\label{ns_random_v_1} Exact v]{\includegraphics[width=0.329\linewidth]{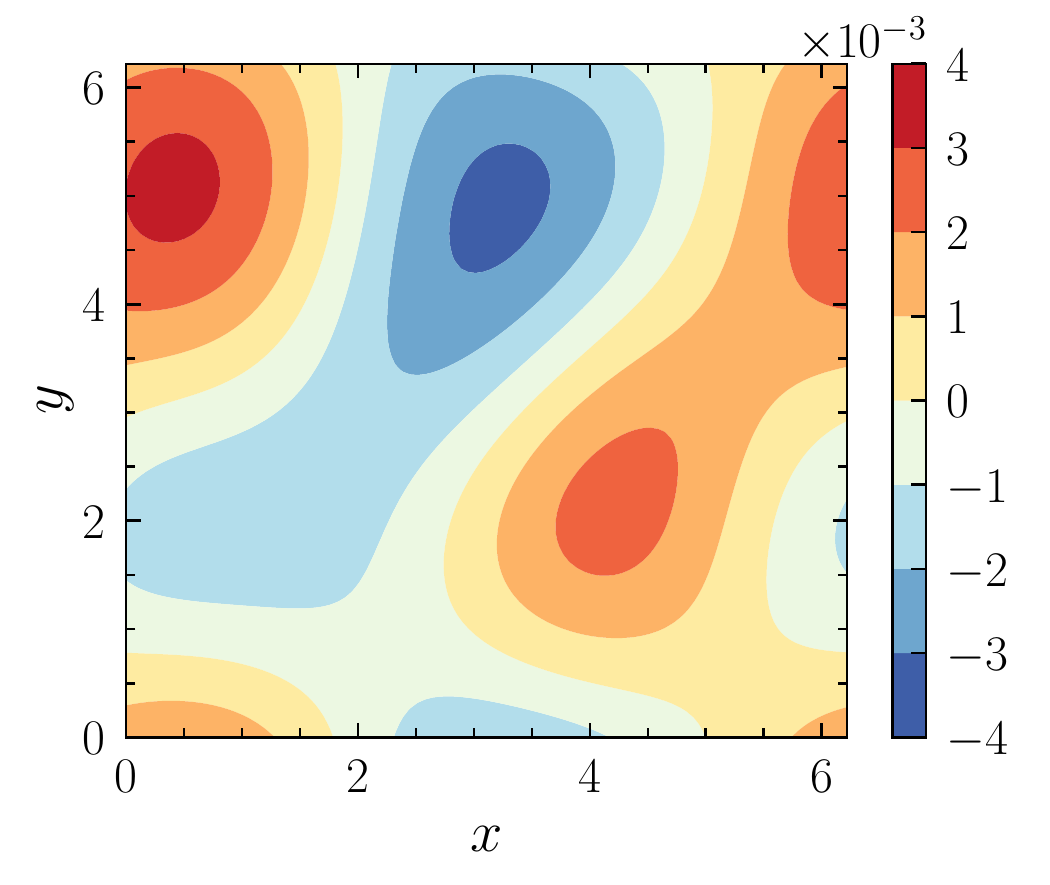}}
   \subfigure[\label{ns_random_v_2} Predicted v]{\includegraphics[width=0.329\linewidth]{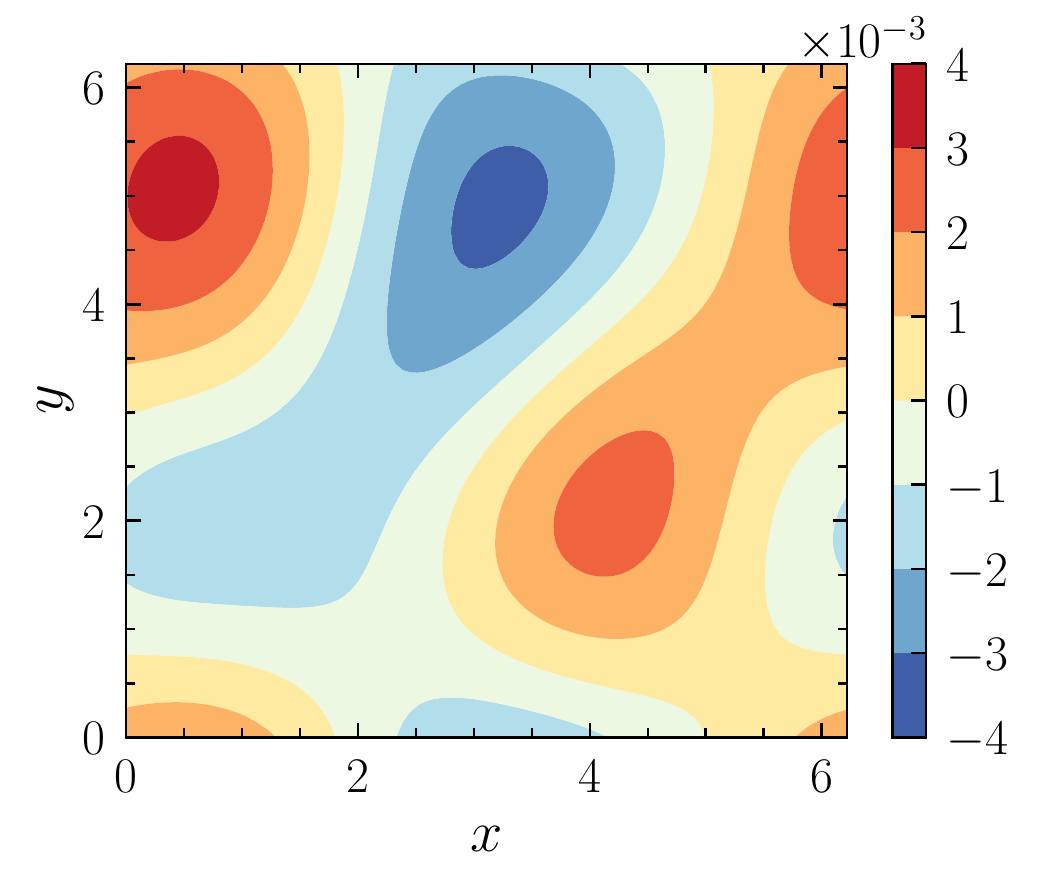}}
   \subfigure[\label{ns_random_v_3} Squared error of v]{\includegraphics[width=0.329\textwidth]{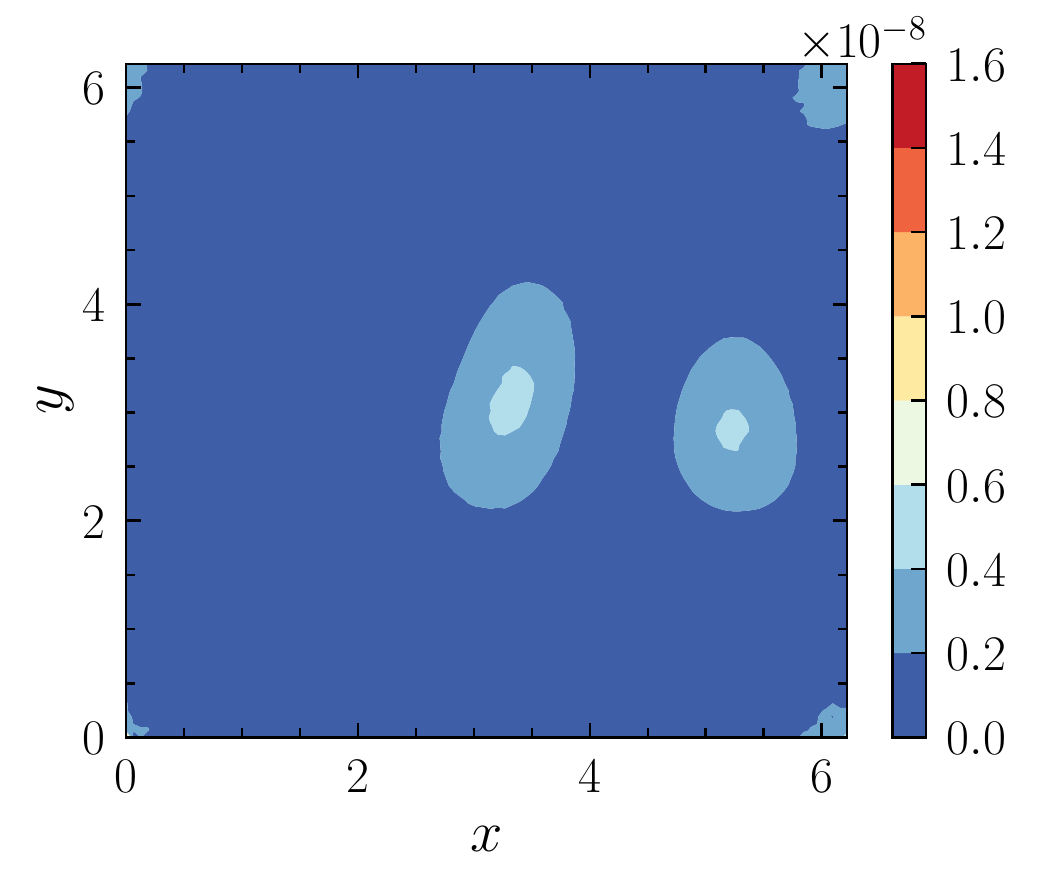}} 
  \caption{For the 2-D NS equations \cref{eq:ns} with the Gaussian random initial condition, exact solutions of $u(2,\boldsymbol{x})$ and $v(2,\boldsymbol{x})$ are plotted on \cref{ns_random_1,ns_random_v_1} respectively, predicted solutions by SINNs of $u(2,\boldsymbol{x})$ and $v(2,\boldsymbol{x})$ are plotted on \cref{ns_random_2,ns_random_v_2} respectively,and the corresponding squared errors are plotted \cref{ns_random_3,ns_random_v_3}}
  \label{fig:NS_random}
\end{figure}

\paragraph{NS equations with random initialization (NS\_random)}
The 2-D NS equation \cref{eq:ns} is also included here with $T=2$, $\nu=2\pi/100$, and a random initial condition $\boldsymbol{\boldsymbol{g}(\boldsymbol{x})}$, namely,
\begin{equation}\label{GaussIni}
\begin{gathered}
        \hat{\boldsymbol{u}}(0,\boldsymbol{k})=\hat{\boldsymbol{g}}(\boldsymbol{k}),\\
        \hat{\boldsymbol{g}}(\boldsymbol{k}) = \left\{ \begin{aligned}
    10^4 \sqrt{ 0.123456 / H(1)} 
    (\boldsymbol{h} - \boldsymbol{k}\boldsymbol{k}\cdot\boldsymbol{h}/|\boldsymbol{k}|^2) & , 
      ~~{\frac{1}{2}} \leq |\boldsymbol{k}| < {\frac{3}{2}} ,\\
    10^4 \sqrt{ 0.654321 / H(2)} 
    (\boldsymbol{h} - \boldsymbol{k}\boldsymbol{k}\cdot\boldsymbol{h}/|\boldsymbol{k}|^2) & , 
      ~~{\frac{3}{2}} \leq |\boldsymbol{k}| < {\frac{5}{2}} ,\\
    10^4 \sqrt{ 0.345612 / H(3)} 
    (\boldsymbol{h} - \boldsymbol{k}\boldsymbol{k}\cdot\boldsymbol{h}/|\boldsymbol{k}|^2) & , 
      ~~{\frac{5}{2}} \leq |\boldsymbol{k}| < {\frac{7}{2}} ,\\
    \boldsymbol{0} & , ~~|\boldsymbol{k}| \geq {\frac{7}{2}} ,
        \end{aligned} \right.\\
        H(n) = \sum\limits_{n-{\frac{1}{2}}\leq |\boldsymbol{k}|< n+{\frac{1}{2}}} |\boldsymbol{h}(\boldsymbol{k})|^2 , 
\end{gathered}
\end{equation}
where $h \in \mathbb{C}$ generated by standard normal distribution fulfills the symmetry $h(\boldsymbol{k})=\bar{h}(-\boldsymbol{k})$.
The spatial and temporal domains are discretized to $100 \times 100$ and $11$ points, respectively.
And the total size for PINNs is $100 \times 100 \times 11$, while the total size for SINNs is reduced to $51 \times 100 \times 11$ for a real $\boldsymbol{u}$.
The predicted solutions from the SINNs for this problem, including the $x$-component $u$ and $y$-component $v$ of the velocity $\boldsymbol{u}(2,\boldsymbol{x})=(u,v)$, are plotted in \cref{fig:NS_random}.
And the corresponding numerical solution $\boldsymbol{u}(2,\boldsymbol{x})$ is obtained with a sufficiently small time step by the spectral method in \cref{Appendix: spectral method}.

\begin{figure}
    \centering
    \includegraphics[width=1\linewidth]{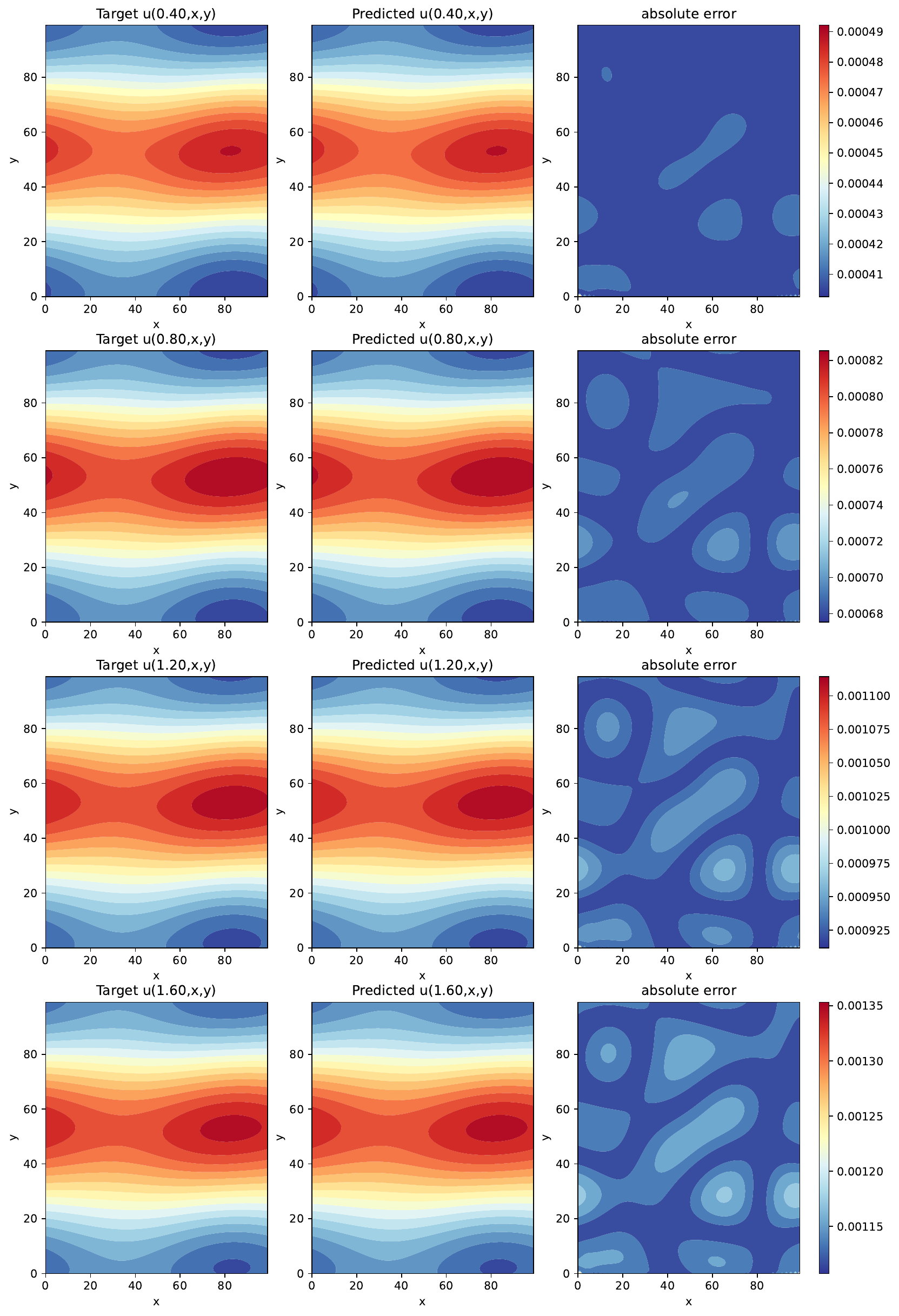}
    \caption{Representative snapshots of the predicted $u$ against the ground truth at $t = 0.4, 0.8, 1.2, 1.6$}
    \label{fig:NS_random_u}
\end{figure}

\begin{figure}
    \centering
    \includegraphics[width=1\linewidth]{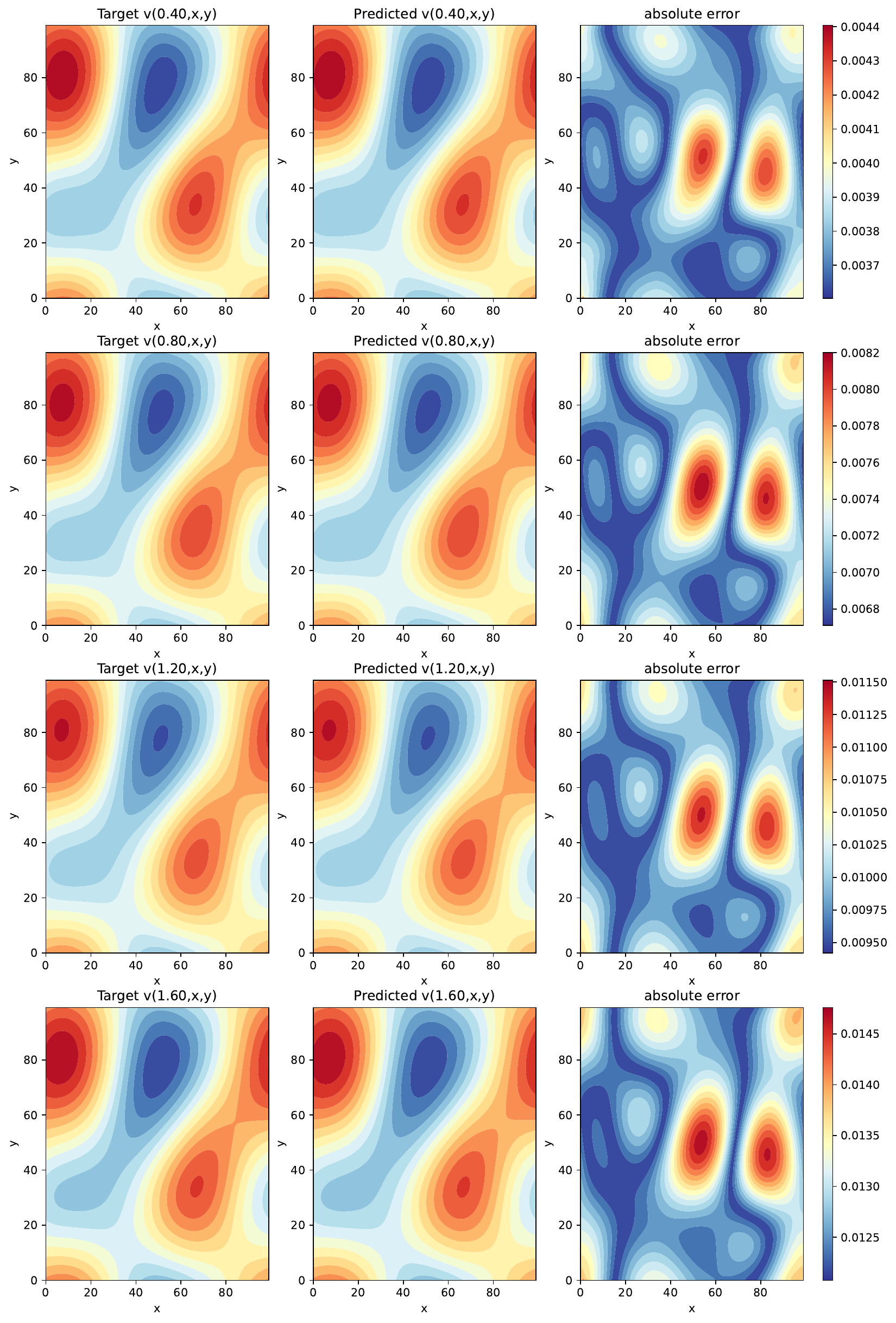}
    \caption{Representative snapshots of the predicted $v$ against the ground truth at $t = 0.4, 0.8, 1.2, 1.6$}
    \label{fig:NS_random_v}
\end{figure}
\subsection{The metrics of the relative error}\label{Relative error}
The metric we use is the relative L2 error as follows:
\begin{equation}
    E=\frac{\sqrt{\sum_{i=1}^N\left|u^\theta\left(t^i,x^i\right)-u^{T}\left( t^i, x^i\right)\right|^2}}{\sqrt{\sum_{i=1}^N\left|u^{T}\left( t^i, x^i\right)\right|^2}},
\end{equation}
where $u^{T}$ is the target solution and $u^\theta$ is the trained approximation. In cases where $u^{T}$ cannot be analytically represented, the spectral method in \cref{Appendix: spectral method} is utilized to obtain high-accuracy numerical solutions.
\section{Some details involved in the spectral form of the incompressible NS equations}\label{Appendix: detailed NS}
This appendix presents the details of the derivation of \cref{PS_NS} and the calculation of non-linear terms $\hat{\boldsymbol{N}}$ in spectral space.

Considering the periodic boundary conditions, by applying the Fourier transform on both sides of \cref{eq:ns}, the NS equations in the spectral space are expressed as
\begin{subequations}
\begin{equation}
    i \boldsymbol{k} \cdot \hat{\boldsymbol{u}} = 0 ,
\end{equation}
\begin{equation}\label{PS_mom2}
\partial_t \hat{\boldsymbol{u}} + \hat{\boldsymbol{N}} = - i \boldsymbol{k}  \hat{p} - \nu |\boldsymbol{k}|^2 \hat{\boldsymbol{u}} .
\end{equation}
\end{subequations}
The continuity equation reveals that frequency and velocity are orthogonal in spectral space; 
 by taking the frequency dot product on both sides of the momentum equation \cref{PS_mom2}, the relationship between the pressure and the non-linear term can be obtained,
\begin{equation}
    \boldsymbol{k} \cdot\hat{\boldsymbol{N}} = - i |\boldsymbol{k}|^2 \hat{p} .
\end{equation}
Eliminating the pressure in the momentum equation \cref{PS_mom2}, the form of \cref{PS_momentum} can be finally obtained.

According to the identities in field theory, the non-linear terms in \cref{eq:ns_momentum} can be expressed in the form
\begin{equation}
    \boldsymbol{N} = \boldsymbol{u}\cdot \boldsymbol{\nabla} \boldsymbol{u} 
      = \boldsymbol{\nabla}\left(\boldsymbol{u}\cdot\boldsymbol{u}/2\right) + (\boldsymbol{\nabla}\times\boldsymbol{u})\times \boldsymbol{u} ,
\end{equation}

  where the term $\boldsymbol{\nabla}\left(\boldsymbol{u}\cdot\boldsymbol{u}/2\right)$ has no contribution to \cref{PS_momentum}, and the non-linear term can be simplified as the rotational form~\cite{Canuto1988Spectral} $\boldsymbol{N} = (\boldsymbol{\nabla}\times\boldsymbol{u})\times \boldsymbol{u}$.
The specific calculation of non-linear terms in spectral space can be written as
\begin{equation}
    \boldsymbol{N} = \mathcal{D}\left[ 
      \mathcal{F}^{-1}\left[ i\boldsymbol{k}\times\hat{\boldsymbol{u}}\left(t,\boldsymbol{k}\right) \right] 
      \times \mathcal{F}^{-1}\left[\hat{\boldsymbol{u}}\left(t,\boldsymbol{k}\right)\right] 
    \right],
\end{equation}
where $\mathcal{D}$ is the dealiasing operator in \cref{Appendix: aliasing}.

\section{Simplification of weighted residual loss}\label{Appendix: Weighted loss}
Suppose 
\begin{equation}
    \mathcal{F}_i=\left|\partial_{t}\hat{u}^\theta\left(t_r^i, \boldsymbol{k}^i\right)
      +\left(k_x^i\right)^2\hat{u}^\theta\left(t_r^i, \boldsymbol{k}^i\right)
      +\left(k_y^i\right)^2\hat{u}^\theta\left(t_r^i, \boldsymbol{k}^i\right)\right|^2,
\end{equation}
and $\boldsymbol{\mathcal{L}}_r=\{\mathcal{F}_i\}_{i=0}^{N_r} $ is the vectorization of $\mathcal{F}_i$ sorted by $\|\boldsymbol{k}\|_1$, then the  \cref{eq:cauloss_c} can be simplified to :
\begin{equation}
    \Tilde{\mathcal{L}}_r(\theta)=\frac{1}{M} \left(\exp{\left(-\epsilon W\boldsymbol{\mathcal{L}}_r\right)}\right)^T\boldsymbol{\mathcal{L}}_r.\label{eq:cauloss}
\end{equation}
where $W=\{w_{ij}\} \in  \mathbb{R}^{N_r\times N_r}$ and 
\begin{equation}
    w_{ij}=\left\{
\begin{aligned}
1 & \quad \text{ if } j < \#\{\|\boldsymbol{k}\|_1\leq \|\boldsymbol{k}^i\|_1\},\\
0 & \quad \text{ otherwise.}
\end{aligned}
\right.
\end{equation}
Furthermore, we can decompose $W=W^1W^2W^3$ where  $W^1=\{w^1_{ij}\} \in  \mathbb{R}^{N_r\times (M-1)}$,$W^2=\{w^2_{ij}\} \in  \mathbb{R}^{(M-1)\times (M-1)}$, and $W^3=\{w^3_{ij}\} \in  \mathbb{R}^{(M-1)\times N_r}$. For 
\begin{align}
& w^1_{ij}=\left\{
\begin{aligned}
1 & \quad \text{ if } \#\{\|\boldsymbol{k}\|_1\leq \|\boldsymbol{k}^i\|_1\}<j\leq \#\{\|\boldsymbol{k}\|_1\leq \|\boldsymbol{k}^i\|_1+1\},\\
0 & \quad \text{ otherwise.}
\end{aligned}
\right.\\
& w^2_{ij}=\left\{
\begin{aligned}
1 & \quad \text{ if } j\leq i,\\
0 & \quad \text{ otherwise.}
\end{aligned}
\right. \\
& w^3_{ij}=\left\{
\begin{aligned}
1 & \quad \text{ if } \#\{\|\boldsymbol{k}\|_1< \|\boldsymbol{k}^j\|_1\}<i\leq \#\{\|\boldsymbol{k}\|_1\leq \|\boldsymbol{k}^j\|_1\},\\
0 & \quad \text{ otherwise.}
\end{aligned}
\right.
\end{align}

\section{Comparison of training time}\label{Appendix: time}

\begin{table}[ht]
\caption{Comparison of the training time of the linear equations}
\label{table:timelinear}
\centering
\begin{tabular}{clccc}
\toprule
 & equation                   & PINN & SINN (SI) &SINN (WL) \\
\cmidrule(r){1-5}
\multirow{2}{*}{1-D}           & convection-diffusion$^\dag$           & 1.92 h  & \textbf{ 1.63 h} &  1.71 h   \\
\cmidrule(r){3-5}
                              & diffusion$^\dag$                         &  \multicolumn{3}{c}{See \cref{table: differenorder}}    \\
                              \cmidrule(r){1-5}
\multirow{2}{*}{2-D}           & heat$^\dag$                              &  8.67 h   & \textbf{5.28 h}  &  7.68 h   \\
\cmidrule(r){3-5}
                              & heat\_random$^\dag$                      & 7.87 h    &  \textbf{4.78 h}  &  7.01 h    \\
                              \cmidrule(r){1-5}
\multirow{1}{*}{3-D}           & heat$^\ddag$                          &  20.94 h  & \textbf{ 14.11 h} &  15.03 h   \\
\bottomrule
\end{tabular}
\end{table}
\begin{table}[ht]
\caption{Comparison of the training time of the non-linear equations}
\label{table:timenonlinear}
\centering
\begin{tabular}{clll}
\toprule
 & equation                   & PINN & SINN \\\cmidrule(r){1-4}
\multirow{1}{*}{1-D}           & Burgers$^\dag$                           & 2.03 h   & \textbf{ 1.51 h}   \\
\cmidrule(r){1-4}
\multirow{2}{*}{2-D}           & NS\_TG$^\ddag$        & 3.38 h   &  \textbf{2.51 h}  \\
\cmidrule(r){3-4}
                              & NS\_random$^\ddag$    & 3.40 h   &  \textbf{2.53 h}    \\
\bottomrule
\end{tabular}
\end{table}

Those superscripts are:

\dag : train on single Tesla V100-SXM2-16GB with CUDA version: 12.3.

\ddag : train on single NVIDIA A100-SXM4-80GB with CUDA version: 11.2.

\section{Ablation experiments}\label{Appendix: ablation}
We did the ablation experiments on diffusion equation~\cref{eq:diffusion} with $N=4$ on the hyperparameter domain $\alpha=\{0,1,2,3,4\},\beta=\{0,1,2,3,4\},\epsilon=\{1,10^{-1},10^{-2},10^{-3},10^{-4},10^{-5},10^{-6}\}$. The best hyperparameter for WL is $\epsilon=10^{-5}$, and the corresponding error is $1.58\times 10^{-6}\pm6.94\times 10^{-7}$ (see ~\cref{table:ablation_WL}); The best hyperparameter for SI is $\alpha=3, \beta=0$, and the corresponding error is $1.73\times 10^{-6}\pm1.41\times 10^{-6}$ (see ~\cref{table:ablation_SI}); The best hyperparameter for combination of SI and WL is $\alpha=1, \beta=0, \epsilon=10^{-5}$, and the corresponding error is $2.14\times 10^{-6}\pm\times 10^{-7}$.
\begin{table}[!htbp]  
\caption{Ablation of WL}  
\label{table:ablation_WL}  
\centering  
\begin{tabular}{cc}  
\toprule  
$\epsilon$& relative error \\  
\midrule  
$1.0$& $1.25\times 10^{-4}\pm6.26\times 10^{-5}$ \\
$10^{-1}$& $8.85\times 10^{-5}\pm5.79\times 10^{-5}$ \\
$10^{-2}$& $1.23\times 10^{-4}\pm7.21\times 10^{-5}$ \\
$10^{-3}$& $4.40\times 10^{-5}\pm1.54\times 10^{-5}$ \\
$10^{-4}$& $7.13\times 10^{-6}\pm9.40\times 10^{-7}$ \\
$10^{-5}$& $1.58\times 10^{-6}\pm6.94\times 10^{-7}$ \\
$10^{-6}$& $8.15\times 10^{-6}\pm4.42\times 10^{-6}$ \\
\bottomrule  
\end{tabular}  
\end{table}
\begin{table}[!htbp]  
\caption{Ablation of SI}  
\label{table:ablation_SI}  
\centering  
\scalebox{0.75}{
\begin{tabular}{ccccc}  
\toprule  
$\alpha /\beta$& 0& 1& 2 &3 \\  
\midrule  
0             & $8.94\times 10^{-6}\pm5.33\times 10^{-6}$ & $2.28\times 10^{-6}\pm1.62\times 10^{-6}$ & $9.63\times 10^{-6}\pm1.04\times 10^{-5}$ & $2.51\times 10^{-6}\pm2.18\times 10^{-6}$ \\
1             & $3.54\times 10^{-6}\pm2.51\times 10^{-6}$ & $5.78\times 10^{-6}\pm2.09\times 10^{-6}$ & $4.94\times 10^{-6}\pm5.26\times 10^{-6}$ & $2.29\times 10^{-6}\pm1.38\times 10^{-6}$ \\  
2             & $3.52\times 10^{-5}\pm3.43\times 10^{-5}$ & $6.43\times 10^{-6}\pm7.40\times 10^{-6}$ & $9.31\times 10^{-6}\pm1.18\times 10^{-5}$ & $3.06\times 10^{-5}\pm2.96\times 10^{-5}$ \\  
3             & $1.73\times 10^{-6}\pm1.41\times 10^{-6}$ & $3.51\times 10^{-6}\pm3.47\times 10^{-6}$ & $9.14\times 10^{-6}\pm8.17\times 10^{-6}$ & $5.74\times 10^{-6}\pm4.07\times 10^{-6}$ \\ 
\bottomrule  
\end{tabular}}
\end{table}
\section{Details of spectral method}\label{Appendix: spectral method}
Since the analytical solutions to the 1-D Burgers equation \cref{eq: burgers} as well as the 2-D heat equation \cref{eq: 2-D heat_random} and NS equation \cref{GaussIni} with random initialization are difficult to obtain, we developed in-house spectral method codes to provide the corresponding numerical solutions instead.

Specifically, the 1-D Burgers equation \cref{eq: burgers} in frequency space is 
\begin{equation}
    \hat{u}_t = - \nu k^2 \hat{u} - \mathcal{F}\left[ u u_x\right], \quad k\in [-N/2,N/2-1], t \in \left[0,T\right].
\end{equation}
And the time derivative $\hat{u}_t$ can be approximated by the optimal third-order total variation diminishing Runge-Kutta scheme~\cite{Gottlieb1998}, which has the following explicit discrete form:
\begin{equation}\begin{gathered}
    \hat{u}_1=\hat{u}(t,k) - \Delta t \left\{ 
        \nu k^2 \hat{u}(t,k) + \mathcal{F}\left[  \mathcal{D}\left[ 
        \mathcal{F}^{-1}\left[ \hat{u}\left(t,k\right) \right] 
        \cdot \mathcal{F}^{-1}\left[ ik\hat{u}\left(t,k\right) \right] 
        \right]  \right]
    \right\} ,\\
    \hat{u}_2={\frac{3}{4}}\hat{u}(t,k) + {\frac{1}{4}}\hat{u}_1 
      - {\frac{1}{4}}\Delta t \left\{ 
        \nu k^2 \hat{u}_1 + \mathcal{F}\left[  \mathcal{D}\left[ 
        \mathcal{F}^{-1}\left[ \hat{u}_1 \right] 
        \cdot \mathcal{F}^{-1}\left[ ik\hat{u}_1 \right] 
        \right]  \right]
    \right\} ,\\
    \hat{u}(t+\Delta t,k) = {\frac{1}{3}}\hat{u}(t,\boldsymbol{k}) + {\frac{2}{3}}\hat{u}_2
      - {\frac{2}{3}}\Delta t \left\{ 
        \nu k^2 \hat{u}(t,k) + \mathcal{F}\left[  \mathcal{D}\left[ 
        \mathcal{F}^{-1}\left[ \hat{u}_2\right] 
        \cdot \mathcal{F}^{-1}\left[ ik\hat{u}_2\right] 
        \right]  \right]
    \right\} ,
\end{gathered}\end{equation}
  where $\Delta t$ is the time step, $\mathcal{D}$ is the dealiasing operator based on the Fourier smoothing method~\cite{Hou2007}, and $\hat{u}_1$ and $\hat{u}_2$ are intermediate variables.

Besides, the 2-D heat equation \cref{eq: 2-D heat_random} can be written in frequency space as 
\begin{equation}
        \hat{u}_t = - \epsilon\left( k_x^2 + k_y^2 \right) \hat{u}, 
        \quad \boldsymbol{k}=(k_x,k_y) \in [-N/2,N/2-1]^2, \ t \in [0,T].
\end{equation}
The optimal third-order total variation diminishing Runge-Kutta scheme~\cite{Gottlieb1998} can be employed for the time derivative $\hat{u}_t$, and the explicit discrete forms for $\hat{u}$ are
\begin{equation}\begin{gathered}
    \hat{u}_1=\hat{u}(t,\boldsymbol{k})-\epsilon\Delta t\left( k_x^2 + k_y^2 \right) \hat{u}(t,\boldsymbol{k}),\\
    \hat{u}_2={\frac{3}{4}}\hat{u}(t,\boldsymbol{k}) + {\frac{1}{4}}\hat{u}_1 
             - {\frac{1}{4}}\epsilon\Delta t\left( k_x^2 + k_y^2 \right) \hat{u}_1 ,\\
    \hat{u}(t+\Delta t,\boldsymbol{k}) = {\frac{1}{3}}\hat{u}(t,\boldsymbol{k}) + {\frac{2}{3}}\hat{u}_2
             - {\frac{2}{3}}\epsilon\Delta t\left( k_x^2 + k_y^2 \right) \hat{u}_2 .
\end{gathered}\end{equation}

As for the 2-D NS equation \cref{GaussIni}, the second-order Adams-Bashforth scheme~\cite{Orszag1971} is applied to the time discretization, and the explicit discrete system becomes
\begin{small}
\begin{equation} 
\begin{gathered}
\hat{\boldsymbol{u}}(t+\Delta t,\boldsymbol{k}) = e^{(-\nu |\boldsymbol{k}|^2 \Delta t)}
  \left(1- \frac{\boldsymbol{k}\boldsymbol{k}\cdot }{|\boldsymbol{k}|^2}\right)
  \left[ 
    {\frac{3\Delta t}{2}}\hat{\boldsymbol{N}}(t,\boldsymbol{k}) 
    - {\frac{\Delta t}{2}}e^{(-\nu |\boldsymbol{k}|^2 \Delta t)}\hat{\boldsymbol{N}}(t-\Delta t,\boldsymbol{k})
    - \hat{\boldsymbol{u}}(t,\boldsymbol{k})
  \right] , \\
\hat{\boldsymbol{N}}(t,\boldsymbol{k})  = \mathcal{F}\left[  \mathcal{D}\left[ 
      \mathcal{F}^{-1}\left[ i\boldsymbol{k}\times\hat{\boldsymbol{u}}\left(t,\boldsymbol{k}\right) \right] 
      \times \mathcal{F}^{-1}\left[\hat{\boldsymbol{u}}\left(t,\boldsymbol{k}\right)\right] 
    \right]  \right], \\
\hat{\boldsymbol{N}}(t-\Delta t,\boldsymbol{k})  = \mathcal{F}\left[  \mathcal{D}\left[ 
      \mathcal{F}^{-1}\left[ i\boldsymbol{k}\times\hat{\boldsymbol{u}}\left(t-\Delta t,\boldsymbol{k}\right) \right] 
      \times \mathcal{F}^{-1}\left[\hat{\boldsymbol{u}}\left(t-\Delta t,\boldsymbol{k}\right)\right] 
    \right]  \right], \\
\end{gathered}
\end{equation}
\end{small}
where $\mathcal{D}$ is the dealiasing operator based on the Fourier smoothing method~\cite{Hou2007}.

In our calculations, the time steps for the aforementioned three discrete forms are chosen to be small enough to minimize the impact of numerical errors on the solutions.
\end{document}